\documentclass[final]{elsarticle}
\usepackage[a4paper]{geometry}
\usepackage[utf8]{inputenc}
\usepackage{booktabs}
\usepackage{bbm}
\usepackage{graphicx}
\usepackage{schemata}
\usepackage{color,soul}
\usepackage{amsmath}
\usepackage{multirow}
\usepackage{amsfonts}
\usepackage[table]{xcolor}
\newtheorem{definition}{Definition}[section]   
\usepackage{makecell}

\usepackage{soul}
\usepackage{units}
\usepackage{tikz}
\usepackage{standalone, times}
\usetikzlibrary{mindmap,backgrounds}
\usetikzlibrary{positioning}

\usepackage{algorithm}
\usepackage{algorithmic}

\usepackage{enumitem}
\usepackage{float}

\usepackage[toc,page]{appendix}
\usepackage[modulo]{lineno}
\usepackage{graphicx}
\usepackage{varioref}
\usepackage{array}
\usepackage{amsmath}
\usepackage{booktabs}
\usepackage{multirow}
\usepackage{subcaption}
\usepackage{array,arydshln}
\usepackage{amssymb}
\usepackage{hhline}
\usepackage{times,rotating}
\usepackage{comment}
\usepackage{tikz}
 
 \tikzset{variable/.default=}
\usepackage{tabularx,ragged2e}
\newcolumntype{T}[1]{>{\raggedright\arraybackslash}p{#1}}
\newcolumntype{M}[1]{>{\centering\arraybackslash}m{#1}}

\newcolumntype{L}[1]{>{\raggedright\let\newline\\\arraybackslash\hspace{0pt}}m{#1}}
\newcolumntype{C}[1]{>{\centering\let\newline\\\arraybackslash\hspace{0pt}}m{#1}}
\newcolumntype{R}[1]{>{\raggedleft\let\newline\\\arraybackslash\hspace{0pt}}m{#1}}

\DeclareMathOperator*{\argmin}{arg\,min}
\DeclareMathOperator*{\argmax}{arg\,max}
\usepackage{multirow}
\usepackage{tikz}
\usepackage[edges]{forest}
\useforestlibrary{linguistics}
\forestapplylibrarydefaults{linguistics}
\usepackage{pifont}
\usepackage{physics}

\usepackage{framed} 
\usepackage{multicol} 
\usepackage{nomencl} 

\usepackage{booktabs}       
\usepackage{amsfonts}       
\usepackage{nicefrac}       
\usepackage{microtype}      
\usepackage{diagbox} 
\usepackage{dsfont}
\usepackage{bm} 
\usepackage{amsmath}
\usepackage{graphicx}
\usepackage{amsfonts}
\usepackage{subcaption}
\usepackage{stmaryrd}
\usepackage{color, colortbl}
\usepackage{breakcites}
\usepackage{pdfpages}
\usepackage{rotating}
\usepackage{dirtree} 

\newcounter{example}[section]
\newenvironment{example}[1][]{\refstepcounter{example}\par\medskip
   \noindent \textbf{Example~\theexample. #1} \rmfamily}{\medskip}

\usepackage[colorinlistoftodos]{todonotes} 
\usepackage[colorlinks=true, allcolors=blue]{hyperref}

\newcommand{\adri}[1]{\textcolor{black}{#1}}
\newcommand{\adrien}[1]{\textcolor{black}{#1}}

\definecolor{orange}{HTML}{FFC17D}
\definecolor{green}{HTML}{A1D68B}
\definecolor{lightgray}{HTML}{E8E8E8}

\usepackage{hyperref} 

\makenomenclature
\renewcommand*\nompreamble{\begin{multicols}{2}}
\renewcommand*\nompostamble{\end{multicols}}

\usetikzlibrary{arrows.meta,shadows.blur}

\begin{document}

\begin{frontmatter}

\title{Greybox XAI: a Neural-Symbolic learning framework to produce interpretable predictions for image classification}

\author[a,b,c]{Adrien Bennetot}
\author[a]{Gianni Franchi}
\author[d,e]{Javier Del Ser}
\author[c]{Raja Chatila}
\author[f]{Natalia D\'iaz-Rodr\'iguez}

\address[a]{U2IS, ENSTA, Institut Polytechnique Paris and Inria Flowers, 91762, Palaiseau, France}
\address[b]{Segula Technologies, Parc d'activit\'e de Pissaloup, Trappes, France}
\address[c]{Sorbonne Universit\'e, Paris, France}
\address[d]{TECNALIA, Basque Research and Technology Alliance (BRTA), 48160 Derio, Bizkaia, Spain}
\address[e]{University of the Basque Country (UPV/EHU), 48013 Bilbao, Spain}
\address[f]{Andalusian Research Institute in Data Science and Computational Intelligence (DaSCI), University of Granada, Spain}

\newpage

\begin{abstract} 

Although Deep Neural Networks (DNNs) have great generalization and prediction capabilities, their functioning does not allow a detailed explanation of their behavior. Opaque deep learning models are increasingly used to make important predictions in critical environments, and the danger is that they make and use predictions that cannot be justified or legitimized. Several eXplainable Artificial Intelligence (XAI) methods that separate explanations from machine learning models have emerged, but have shortcomings in faithfulness to the model actual functioning and robustness. As a result, there is a widespread agreement on the importance of endowing Deep Learning models with explanatory capabilities so that they can themselves provide an answer to why a particular prediction was made. First, we address the problem of the lack of universal criteria for XAI by formalizing what an explanation is. We also introduced a set of axioms and definitions to clarify XAI from a mathematical perspective. Finally, we present the \textit{Greybox XAI}, a framework that composes a DNN and a transparent model thanks to the use of a symbolic Knowledge Base (KB). We extract a KB from the dataset and use it to train a transparent model (i.e., a logistic regression). An encoder-decoder architecture is trained on RGB images to produce an output similar to the KB used by the transparent model. Once the two models are trained independently, they are used compositionally to form an explainable predictive model. We show how this new architecture is accurate and explainable in several datasets. 

\end{abstract}


\end{frontmatter}

\section{Introduction}

Deep Neural Networks (DNNs), considered as black-boxes, are being increasingly used to make important predictions in critical contexts. At the same time, the demand for transparency from the various stakeholders in AI \cite{Preece18Stakeholders} is also growing. The danger of using black box models would be to create and apply decisions that are not explainable and could not be justified \cite{gunning2017explainable}. Common needs for requiring interpretability in Deep Learning are the need for ethics \cite{goodman2017european}, safety when using AI in critical settings \cite{caruana2015Transferability} and the need to enable the end user to trust the system \cite{ExplainableAIForDesigners}. Interpretability can be defined as the ability to produce explanations of the model's behavior that can be understood by a human user \cite{arrieta2020explainable}. 

In \cite{Miller19}, Miller highlighted four major findings about explanations. A prerequisite for a "good" explanation is that it does not only indicate why the model made a certain decision, but also why it made this decision rather than another. This refers to the ability to produce counterfactuals. In addition, "good" explanations are selective, meaning that focusing solely on the main causes of a decision-making process is sufficient. Furthermore, \adri{if a causal explanation for the generalization itself is not provided, utilizing statistical generalizations to explain why occurrences occur is insufficient}. Finally, explanations are social, meaning that they are a transfer of knowledge between an explainer and an explainee. 

Deep Learning models suffer from two kind of bias. The first one is a learning bias, \adri{due to the presence of a bias in the training data set}. This can happen when associations of concepts are over- or under-represented in the training set. For example there were cases of a dataset with women under-represented in offices compared to men, leading a captioning algorithm to assume that a person in an office was necessarily a man while it could also be a woman. One of the applications of XAI is highlighting this bias \adri{to correct it} \cite{hendricks2018women}. The second type of bias is the human induced one, when using common sense knowledge about the world to explain the output of a DNN or when using particular parameters, architectures or loss functions to model a problem \cite{Doran17}. 

A large number of methods for model probing have emerged in recent years. Some have the advantage of being model-agnostic, i.e. separating the explanation from the machine learning model. This has the advantage of providing flexibility to the user as tools are available to extract explanatory elements from each model \cite{LIME}. Some of these methods, the best known of which are LIME \cite{LIME1} and SHAP \cite{lundberg2017unified}, are based on the use of surrogate models. These proxy models will locally mimic the behaviour of the black-box in order to explain individual predictions. While this has the advantage of being easy to use, there are problems of robustness \cite{Alvarez18, Slack19}. Moreover, it is not possible to have a global view of the model's behaviour since the explanation is local. 

In image recognition, another family of methods widely used is based on visualization. They express an explanation by highlighting characteristics of the image that objectively influence the output of a DNN \cite{ras2021explainable}. The best known of them, Grad-CAM \cite{selvaraju2017grad}, creates class activation map using the gradients of the DNN's output with respect to the last convolutional layer. This provides a visual explanation easy to understand as it allows recognizing the important regions of the image. However it is difficult to know whether an explanation is \adri{valid}, \adri{in the sense that a human non-expert in the field does not necessarily know what the important points of an image are}, and a part of the evaluation is subjective. Furthermore, it has been shown that some of the most used methods are insensitive to model and data \cite{adebayo2018sanity}. In addition, there is also a risk of introducing a human-induced bias when a user is trying to interpret the visual explanation. His \adri{or her} understanding would depend on his \adri{or her} own background knowledge. Thus, it is necessary that the explanatory elements of an AI model come directly from the data seen by the network, for it to be faithful with respect to what it actually learned \cite{Bennetot19}. 

One of the goals of having interpretability in a model is to explain its reasoning by expressing it in a way that is understandable and readable by human beings, while highlighting the biases learned by the model, in order to validate or invalidate its decision rationale \cite{Guidotti19}. There is a trade-off between the performance of a model and its transparency \cite{Dosilovic18} but it is also possible to consider that the advocacy for interpretability may lead to a generic performance improvement for 3 reasons: i) it will help ensure impartiality in decision-making, i.e. to highlight, and consequently, correct from bias in the training data-set, ii) interpretability facilitates the provision of robustness by highlighting potential adversarial perturbations that could change the prediction, and finally, iii) interpretability can act as an insurance that only meaningful variables infer the output, i.e., guaranteeing that an underlying truthful causality exists in the model reasoning. Combining the prediction capabilities of connectionist models with the transparency of symbolic ones could put aside the trade-off by increasing either the interpretability or the performance of AI models, the challenge being to increase one without sacrificing too much of the other. It has been proven that using background knowledge within a DNN can bring robustness to the learning system \cite{Donadello17,donadello2018semantic,dAvilaGarcez19NeSy}. The use of a Knowledge Base to learn and reason with symbolic representations has the advantage of promoting the production of explanations while making a prediction \cite{Donadello19}. The ability to refer to established reasoning rules allows symbolic methods to fulfill this property. 

In order to obtain a model that meets the above criteria, we introduce the \textit{Greybox XAI framework}. This new architecture is transparent by design when used for an image classification task. It combines an encoder-decoder used for the creation of an \textit{Explainable Latent Space} which is then used by a logistic regression. The \textit{Explainable Latent Space} allows knowing for which reasons an image has been classified in a certain way with the help of logistic regression. Moreover, we propose a formalization of the notion of explanation and we pose definitions allowing to judge its quality.

The contribution of our paper is threefold:

\begin{itemize}
    \item A theory of explainability of deep learning models to qualify what is a "good" explanation.
    \item An explainable by design compositional framework called \textit{Greybox XAI}.
    \item We show that this new framework provides state-of-the-art results on an image classification task regarding the explainability/accuracy trade-off in various datasets, as its accuracy is close to the existing models while being more explainable.
\end{itemize}

This paper is organized as follows: first we present the literature around XAI and part-based classifiers in Section \ref{rw}. We present the different notions and terminology used in XAI and propose our definitions in Section \ref{XAI}. We describe our framework in Section \ref{Greybox XAI XAI} and we illustrate its use by experiments on several datasets in Section \ref{Experiments}.

\section{Related Work: Explainable AI formalization, compositional part-based classification and neural-symbolic computation.} \label{rw}




The literature \cite{arrieta2020explainable, guidotti2018survey, buhrmester2019analysis} distinguishes Deep Learning's XAI methods into two categories: transparent models and opaque models that need to be explained thanks to post-hoc methods. As our model is a composition of a transparent model and an opaque model, we will put a particular focus on compositional models in Section \ref{rw:compositional}. We will also talk about the use of KBs for XAI in Section \ref{rw:kb}.

\subsection{Compositional Part-based Classification Models}  \label{rw:compositional}

\adri{Compositionality in computer vision refers to the ability to represent complex concepts by combining simpler parts \cite{andreas2019measuring,  AFodor2002-AFOCP-2}. Compositionality is a desirable property for CNNs as it can improve generalization by encouraging networks to form representations that disentangle the prediction of objects from their surroundings and from each other \cite{stone2017teaching}. For example, handwritten symbols can be learned from only a few examples using a compositional representation of strokes \cite{lake2015human}. The compositionality of neural networks is also seen as key to the integration of symbolism and connectionism \cite{hupkes2019compositionality, mao2019neuro}.}

Part-based object recognition is an example of semantic compositionality and a classical paradigm where the idea is to collect information at the local level in order to make a global classification. In \cite{de1999object}, the authors propose a pipeline that first groups pixels into superpixels, then performs a superpixel-level segmentation, converts this segmentation into a feature vector, and finally classifies the global image thanks to this feature vector. A similar method is proposed by \cite{huber2004parts}, extending it to 3D data. Here, the idea is to classify a part of the image into a predefined class and then use these intermediate predictions to create a classification of the whole image. The authors of \cite{bernstein2005part} also define intermediate-level features that capture local structures such as vertical or horizontal edges, hair filters, and so on. However, they are closer to dictionary learning than to the approach we propose in this paper.

One of the best known models for object part recognition is \cite{felzenszwalb2009object}. It provides object recognition based on mixtures of deformable part models with multiple scales based on data mining of hard negative examples with partially labeled data to train a latent SVM. The evaluation is performed in the PASCAL object detection challenge (PASCAL VOC benchmark \cite{pascal-voc-2012}).

Semi-supervised methods have been developed more recently, such as \cite{ge2019weakly}. They propose a two-stage neural architecture for fine-grained image classification supported by local detections. The idea is that positive proposal regions highlight different complementary information and that all of this information should be used. For this purpose, an unsupervised recognition model is first built by alternately applying a CRF and a Mask-RCNN (considering an initial approximation with CAM). Then, the recognition model and the positive region proposal are fed to a bidirectional LSTM, which generates a meaningful feature vector that collects information about all regions and is then able to classify the image. This can be considered as unsupervised part-based classification.

\adri{As introduced in \cite{Holzinger2019_Causability}, there is a need for causability in certain domains such as in the medical field for example. Causability is the measurable extent to which an explanation to a human expert achieves a specified level of causal understanding \cite{HOLZINGER202128}. This notion refers to usability and must not be confused with causality. The latter is the relationship between cause and effect \cite{pearl2009causality}. Causability can be measured with the System Causability Scale, a system to measure the quality of explanations based on causability and usability \cite{Holzinger_2020_scs}.}

\adri{Current methods for image classification with DNNs is to add attention mechanisms into the learning process to automatically extract relevant features of a given input. This mechanism is designed to focus DNNs on the most important features for a given classification task.\cite{Hu18}. Transformer architecture was first introduced for machine translation but the computer vision community is working on their implementation for computer vision \cite{steiner2021augreg, tolstikhin2021mixer, zhuang2022gsam}. Attention-based neural networks such as the Vision Transformer (ViT) have recently attained state-of-the-art results, in term of accuracy, on many computer vision benchmarks \cite{dosovitskiy2020vit, chen2021outperform}. Many researches have emerged to improve these transformers for computer vision, notably working on techniques to more efficiently scale the size of vision transformers\cite{Zhai_2022_CVPR}, to make them more tractable for inference \cite{Chavan_2022_CVPR} or to improve generalization to domain shift \cite{Zhang_2022_CVPR}. By adding human visual attention maps are as an input for a DNN, it has been shown that attention mechanisms in DNNs work as human visual attention\cite{Obeso2022}.
}

Finally, \cite{Diaz-Rodriguez21} proposes a methodology designed to learn both symbolic and deep representations. It involves a compositional convolutional neural network that makes use of symbolic representations called EXPLANet and SHAP-Backprop, an explainable AI-informed training procedure that corrects and guides the DL process to align with such symbolic representations in form of knowledge graphs. To the best of our knowledge, this model represents the state of the art in terms of compositional learning models.

\subsection{The use of Knowledge base for Explainable AI} \label{rw:kb}

The use of background knowledge in the form of logical statements in KBs has shown to not only improve explainability but also performance with respect to purely data-driven approaches \cite{Donadello17,dAvilaGarcez19NeSy}. A positive side effect shown is that this hybrid approach provides robustness to the learning system when errors are present in the training data labels. Other approaches have shown to be able to jointly learn and reason with both symbolic and sub-symbolic representations and inference \cite{garnelo2019reconciling}. The interesting aspect is that this blend allows for expressive probabilistic-logical reasoning in an end-to-end fashion \cite{manhaeve2018deepproblog}. An example of use case is on dietary recommendations, where explanations are extracted from the reasoning behind (non deep but KB-based) models \cite{Donadello19}.


A different perspective on hybrid XAI models consists of enriching black-box models knowledge with that one of transparent ones, as proposed in \cite{Doran17} and further refined in \cite{Bennetot19}. It allows the network to express what is confident or confused about, in a context that helps to tackle bias. Other examples of hybrid symbolic and sub-symbolic methods where a knowledge-based tool or graph-perspective enhances the neural (e.g., language \cite{petroni2019language}) model are in \cite{Bollacker19,Shang19}.

Another hybrid approach consists of mapping an uninterpretable black-box system to a white-box \textit{twin} that is more interpretable. For example, an opaque Artificial Neural Network (ANN) can be combined with a transparent Case Based Reasoning (CBR) system \cite{Aamodt94, Caruana99}. In \cite{Keane19}, the ANN (in this case a DNN) and the CBR (in this case a k-NN) are paired in order to improve interpretability while keeping the same accuracy. The \textit{explanation by example} consists of analyzing the feature weights of the ANN which are then used in the CBR, in order to retrieve nearest-neighbor cases to explain the ANN’s prediction.

Description Logics \cite{Baader07} have successfully been used for enhancing deep learning models for image interpretation through the use of knowledge bases \cite{donadello2016integration}. It can also help detect inconsistencies in automated knowledge representation and reasoning. An example for automated symbol design and interpretation is in \cite{Lamy17}. Some XAI systems consider counterfactual rule learning and causal signal extractions. Examples of rule learning approaches can include learning from noisy or unstructured data, or learning with constraints \cite{marra2019lyrics}.

As it seems intuitive that the presence of a KB is useful to provide an explanation, how to use it for image classification is not obvious because KBs use a very concrete formalism which is in opposition to the abstract features used by networks. \adri{Some methods such as Logic Tensor Networks \cite{donadello2018semantic} or LYRICS, a General Interface Layer to Integrate AI and Deep Learning \cite{marra2019integrating} show promising results but we did not find any executable implementation example of their end-to-end use in the literature.}



\section{XAI definitions and formalization} \label{XAI}
We first establish a common point of understanding on what the different terms stands for in the context of XAI, based on \cite{arrieta2020explainable}. We recall the definition of transparency and propose various criteria to judge the quality of an explanation. Definitions of subsections \ref{trans} and \ref{expla} are not contributions but a necessary step to properly understand our proposed \textit{Greybox XAI} framework.

\subsection{Definition of transparency}\label{trans}

Transparency refers to a passive property of a model, which refers to the level at which a given model \adri{"makes sense"} to a human observer. A model is considered opaque when it is not transparent. There are 3 degrees of transparency, from the least to the most transparent \cite{Lipton18}: 

\begin{itemize}
    \item Algorithmic Transparency. It deals with the ability of the user to understand the process followed by the model to produce any given output from its input data. 
    \item Decomposability. A model with algorithmic transparency is decomposable if every part of the model is understandable by a human without the need for additional tools. It means that it is possible to explain each of the parts of a model (input, parameter and calculation). It requires every input to be interpretable.
    \item Simulatability. It denotes the ability of a model to be totally simulated by a human, meaning that its complexity is low. A decomposable model is therefore simulatable if it self-contained enough for a human to think and reason about it as a whole. 
\end{itemize}

To conclude, these levels of transparency depend on the understandability of the model. The understandability of a model is its ability to make a human understand its internal structure or the algorithmic means by which the model processes data internally \cite{Montavon18}. The type of models considered as transparent are Linear and Logistic Regression \cite{PurposeLR}, Decision Trees \cite{rokach2008data}, K-Nearest Neighbors \cite{KNNimandoust2013application}, Rule Based Learners \cite{ProductionRulesFromTrees}, General Additive Models \cite{Bankruptcy} and Bayesian Models \cite{BayesianCognitive}. These 6 types of models are always algorithmically transparent. They can reach higher levels of interpretability if the variables are understandable and not too numerous, if there are not too many rules, etc.


\subsection{Clarifying the concept of Explainability}\label{expla}

As stated in \cite{arrieta2020explainable}, explainability can be considered as an active characteristic of a model, which refers to any action or procedure implemented in order to clarify its internal functions. The explainability of a model thus denotes its capacity to produce an explanation. In order to make a non-transparent model explainable, many post-hoc methods were designed. Post-hoc methods are used on a model after its training and are designed to probe the model in order to improve its explainability.

While most of the different terms used in explainable AI have been widely debated in the literature, the one about \textit{What} constitutes an explanation has not been widely mathematized. We propose a formalization of the notion of explanation, inspired by \cite{Alvarez-Melis18} in order to establish objective criteria to affirm that an explanation is \adri{"good"} or not. Whether it comes from the transparency of a model or from a post-hoc method applied on an opaque model, an explanation must meet certain indispensable characteristics to be considered as a \adri{"good"} explanation.

\subsection{Explanation Formalisation for an image classification problem}\label{formalisation}

Let us denote $E = \{(e) | e \in \{0,1\}^*\}$ a set of explanations $e$. This set is binary in order to be able to encode any type of communication because an explanation can be given in various forms: a text, an image, a graph, etc. 
\begin{definition}
Let $f : \mathcal{X} \rightarrow \mathcal{Y}$ be a classification model with $\mathcal{X}$ the input space and $\mathcal{Y}$ the label space. The explanation function $\Phi$ on a $f(x)$ prediction with $x \in \mathcal{X}$ is defined by:

\begin{align}
    \Phi  : \mathcal{Y} &\rightarrow E \\
    f(x) &\rightarrow \Phi (f(x))
\end{align}
\end{definition}
On the basis of this definition of an explanation function, we define several axioms characterizing an explanation. These axioms intend to formally qualify what a \adri{"good"} explanation is, following desired properties. Based on the literature \cite{arrieta2020explainable, Doran17, Miller19} and the above definitions, we highlight 3 properties that we consider necessary to obtain a \adri{"good"} explanation:
\begin{enumerate}
    \item \textbf{Objectivity}. \adrien{Explainability is firstly about humans, as it refers to the details and reasons a model elicits to make its functioning clear or easy to understand given a certain audience \cite{arrieta2020explainable}. The point of producing explanations that are as objective as possible is to minimize the amount of subjectivity a human might have when interpreting the explanations. Designing explanations based on symbols/concepts that are known and related to the task at hand allows for an unbiased explanation that can be understood in the same way by two different users with the same background knowledge.} In order for an explanation to be \adri{more} objective, we consider that it must be expressed in such a way that it is understood in the same way by the majority of members of a given audience. The real world contains objects and we want compact representations of those objects \cite{Baum2004}. We assume this can be obtained in an explanation by the use of logical semantics, using symbols and relations that can be conceptualized by the human user of the explanation. This implies the use of ontologies, specifying what individuals (things, objects) and relationships are assumed to exist and what terminology is used for them. 
    
    \begin{definition}
        An explanation $e$ of a classification model $f : \mathcal{X} \rightarrow \mathcal{Y}$ is said to be objective if $e$ does contain symbols and/or relationships.
    \end{definition}
    
\begin{example}[Objectivity]
Figure \ref{fig:grad_cam} shows the example of one explanation making use of symbols and/or relationships and another one not making use of them. The \adri{less subjective} explanation minimizes the amount of interpretation left to the explainee because it uses symbols (words) commonly employed to represent objects. Moreover the subjective explanation based on a visualization of the attention areas of the model leaves a lot to the explainee's interpretation. From one user to another, some will say that the hot area is the head of the rabbit while others will talk about the color of its muzzle, the carrot strand it is holding or its eyes. 

    \begin{figure}[htbp!]
    \centering
    \includegraphics[scale=0.43]{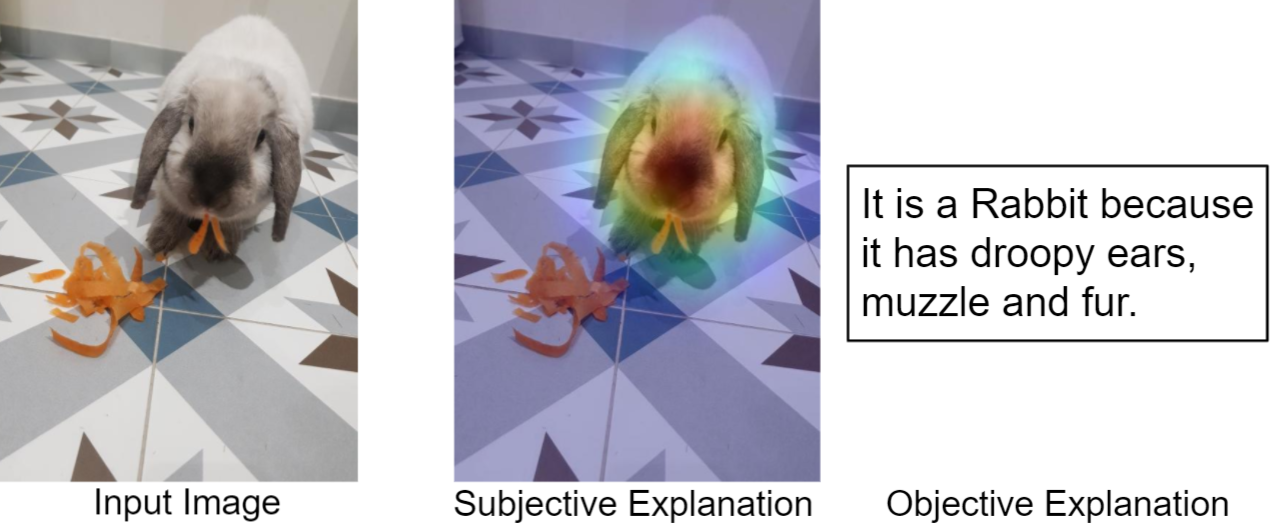} 
    \caption{Subjective and \adri{less subjective/more objective} example of explanations. The subjective explanation is a superimposed visualization of Grad-CAM's heatmap and the input image, showing the model mostly used the center-right of the image (where the head of the rabbit is) in order to make its prediction. The \adri{more} objective explanation is a textual explanation using attributes detected on the rabbit to categorize and describe it.}
    \label{fig:grad_cam}
\end{figure}

\end{example}

    
    \item \textbf{Intrinsicality}. The complete explanation of a prediction should come directly from the model (or its intrinsic elements) that produced the prediction. In order for the explanation to be totally faithful to what happened in the model, it is necessary that only the inputs, parameters and operations present in the model that we are trying to explain are used. This is essential to ensure that the explanation that is given is what actually happened in the model during its inference rather than the expected behavior. The value of having an intrinsic explanation is to be sure that the explanation exactly describes how the model works, rather than an approximate or desired operation. As a matter of fact, if the explanation depends on something that is not related to the model we wish to explain, it is impossible to ensure that this explanation does not distort the real reasons for which a decision was taken. 
    
    \begin{definition}
    An explanation $e$ of a classification model $f : \mathcal{X} \rightarrow \mathcal{Y}$ is said to be intrinsic if $e$ only depend of elements, parameters and operations present in $f$, $\mathcal{X}$ or $\mathcal{Y}$.
    \end{definition}
    
    \begin{example}[Intrinsicality]

    In order to produce an explanation for a black box model $f$, the post-hoc method LIME \cite{LIME1} generates a new dataset consisting of perturbed samples and the corresponding predictions of $f$. On this new dataset, LIME trains an interpretable model $h$, which is weighted by the proximity of the perturbed samples to the instance of interest. The prediction of the model $h$ should be a good approximation of the predictions of the model $f$ locally, but it does not have to be a good global approximation of the model $f$. The produced explanation can be expressed as follows:
    
    \begin{equation}
        e = \Phi(f(x)) = \argmin_h \mathcal{L}(f,h,\pi_x) + \Omega (h)
    \end{equation}
    
    with $\mathcal{L}(f,h,\pi_x)$ the local fidelity, i.e. how close the predictions from $h$ are close to the predictions from $f$. The proximity measure $\pi_x$ defines how large is the neighborhood around the explained instance. Therefore, explanation $e$ does not only depend of elements, parameters and operations present in $f$, $\mathcal{X}$ or $\mathcal{Y}$ since the explanation depends on the surrogate model $h$. Consequently, this explanation is not intrinsic. It is the prediction of the model $h$ that is explained, not that of the model $f$. 
    \end{example}

    \begin{example}[Intrinsicality]
    In opposition, the explanation resulting from a linear regression $h$ can be considered as intrinsic because the learned relationships \adri{between the inputs and the labels} can be written as follows:
    
    \begin{equation}
        e = \Phi(f(x_i)) = \theta_f \times x_i
    \end{equation}
    with $\theta_f$ the set of trainable parameters of $f$ and $x_i$ an instance.
    \end{example}

    \item \textbf{Validity \adri{and Completeness}}. An explanation of a prediction must be valid, meaning that it should assert that the model is Right (or wrong) for the Right Reason (RRR). It must show that the functioning of the model is consistent, that it is not biased by the training data. The explanation should be similar to what an expert in the field would give. To this notion of \textit{validity} we could add a desideratum notion of \textit{completeness}: an explanation could be judged as incomplete if it does not contain enough valid elements in its constitution, \adri{meaning that we want a sufficient number of discriminating elements.} However, it has been established in the literature that a "good" explanation is selective \cite{Miller19}. Selectivity means that humans are adept at selecting \adri{a few} causes from a sometimes infinite number of causes.
    
    
    \begin{definition}
    Given a field expert human being $h : \mathcal{X} \rightarrow \mathcal{Y}$, we define $E_{valid}$ as the set of valid explanations $\Phi : \mathcal{Y} \rightarrow E_{valid}$
    \end{definition}

    \begin{definition}
    An explanation $e$ of a classification model $f : \mathcal{X} \rightarrow \mathcal{Y}$ is said to be valid if $e \in E_{valid}$ \adri{ and complete if $e$ is discriminative enough.}
    \end{definition}
    
    \begin{example}[Validity:]
    
    To illustrate this, we take the example of the image \ref{fig:valid_explanation} of a ram rabbit that we see in its entirety, from the front, and that the model classifies as a rabbit. We can say that the explanation is valid if it contains elements explaining why it is rabbit, i.e. the ones that an expert looking at the image would use to justify the fact that it is a ram rabbit. 
    
    \begin{figure}[htbp!]
    \centering
    \includegraphics[scale=0.41]{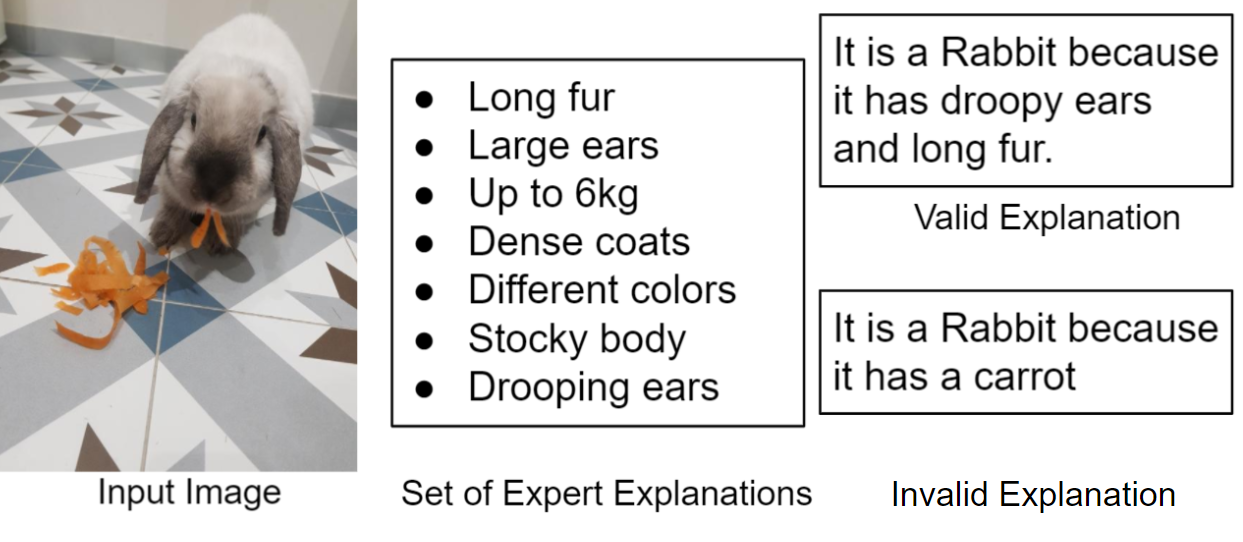} 
    \caption{Valid and invalid examples of explanations. The valid explanation contains elements that are present in the expert explanation set and that are therefore "good" reasons to justify why a rabbit is present in the picture. On the contrary, the invalid explanation contains elements out of the expert explanation set.}
    \label{fig:valid_explanation}
    \end{figure}
    
    \end{example}

    \begin{example}[Completeness]
    
    \adri{To illustrate the notion of completeness, closely related to validity, we take the example on \ref{fig:complete} of a ram rabbit that we see in its entirety, from the front, and that the model classifies as a rabbit. We can say that the explanation is more complete if it contains enough elements explaining why it is a rabbit, i.e., some of the most important features that a human expert looking at the image would notice and use to justify the fact that it is a ram rabbit. Technically, it is not "wrong" to say that a rabbit can be of different colors. However, colour is not a discriminating element to recognize a ram rabbit and it would be improbable to see a human giving this explanation. If this justification on the colors had been accompanied by discriminative elements of the rabbit, such as its big droopy ears, the explanation would have been more complete.
}
    
    \begin{figure}[htbp!]
    \centering
    \includegraphics[scale=0.41]{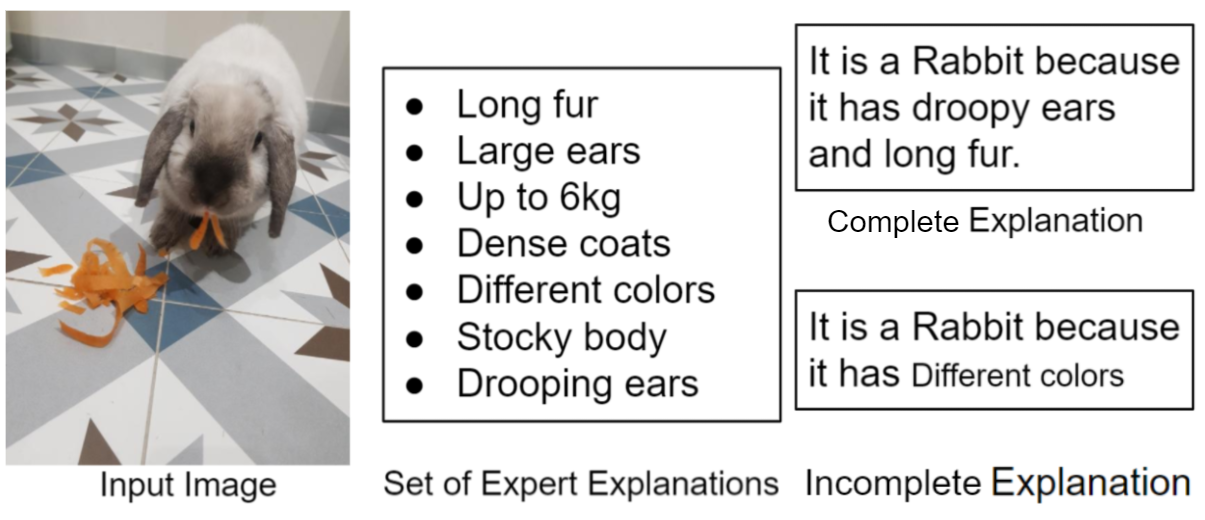}
    \caption{Valid and invalid examples of explanations. The valid explanation contains elements that are present in the expert explanation set and that are therefore "good" reasons to justify why a rabbit is present in the picture. On the contrary, the invalid explanation contains elements out of the expert explanation set.}
    \label{fig:complete}
    \end{figure}
    \end{example}

\end{enumerate}

\section{Greybox XAI: Greybox eXplainable Artificial Intelligence framework} \label{Greybox XAI XAI}

In this section we present the \textit{Greybox XAI framework}. It is designed to be transparent according to the definition of transparency proposed above. This framework is also made to produce "good" explanations in relation to the 3 criteria of objectivity, intrinsicality and validity. The goal of this framework is to perform compositional image classification and to explain its predictions by the different \textit{parts-of} the object that has been classified. It consists of two separately trained models:

\begin{itemize}
    \item A Deep Neural Network trained to predict a segmentation map from an RGB image input. Its purpose is to detect the different \textit{parts-of} objects that constitute the image.
    \item A transparent model trained to predict an object, using as input a vector encoding the presence and absence of \textit{parts-of} objects.
\end{itemize}

These two models are linked in a sequential manner: the output of the DNN is transformed into a vector serving as input to the transparent model. We call the space in which the transformation is carried out the \textit{Explainable Latent Space}. In this space, the predicted segmentation map is transformed into a one-hot vector. This vector indicates all the \textit{parts-of} objects present in the segmentation map. The transparent model classifies this vector. It gives a prediction of the object present on the RGB image according to the different \textit{parts-of}. It is then possible to produce an explanation of this classification based on the \textit{Explainable Latent Space} and the transparent model computation. In the rest of the article, \textit{parts-of} object are called attributes and \textit{objects} are called classes.

First, in Section \ref{Architecture} we describe the architecture of our framework and the required data to exploit its full potential. Then, in Section \ref{TrainingProcess} we explain how we sequentially train the different parts of the framework. Finally, we prove in Section \ref{Inference} how this \textit{Greybox XAI framework} is explainable with regard to the definitions proposed in Section \ref{XAI}.

\subsection{Greybox Architecture and Data Requirements}  \label{Architecture}


Let us denote $X$ and $Y$ two random variables, with $X \sim P_X$ and $Y \sim P_Y$. Without loss of generality we consider the observed samples $\{x_i\}^N_{i=1} \in \mathcal{X}^N$ as vectors and the corresponding labels $\{y_i\}^N_{i=1}\in \mathcal{Y}^N$ as scalars. 
From the set of observations $\mathcal{X}^N$ and the set of corresponding labels $\mathcal{Y}^N$ we derive a training set denoted $\mathcal{D} = \{(x_i, y_i)\}^N_{i=1}$ with N the number of pair elements $(x_i, y_i)$ of dataset $\mathcal{D}$. The elements of dataset $\mathcal{D}$ are assumed to be independent and identically distributed (i.i.d.) according to an unknown joint distribution $P_{X, Y}$. 

Let us denote $f$ a DNN. 
We assume that a DNN is a function that takes two inputs. The first input is the input data $x_i$ and the second input is the set of trainable weights $\theta = \{\theta_k\}_{k=1}^K$ with $K$ the number of weights of the DNN. Hence we denote $f(x_i, \theta)$ the DNN $f$ applied on $x_i$ with the set of weights $\theta$. 

We can consider that a DNN has a probabilistic representation \cite{Blundell2015}; hence a DNN outputs a likelihood probability function of the random variable $Y$ given $X$ parametrized by $\theta$ : $f(x, \theta) = P(y|x, \theta)$.

Using the fact that the training data is independent and identically distributed according to $P_{XY}$, 
the set of training weights $\theta$ are optimized by Maximum Likelihood Estimation (MLE) over the training data $\mathcal{D}$. 

\begin{align} 
    \theta^{MLE} &= \argmax_\theta P(\mathcal{D} | \theta) \\&= \argmax_\theta \prod^N_{i=1} P(y_i | x_i, \theta) \\&= \argmax_\theta \sum^N_{i=1} \log P(y_i | x_i, \theta)
\end{align}

Since the $\argmax$ of a function does not change if we multiply it by a strictly positive scalar, it is possible to write:

\begin{equation}
   \theta^{MLE} = \argmax_\theta \frac{1}{N}\sum^N_{i=1} \log P(y_i | x_i, \theta)
\end{equation}

in order to get to the definition of Cross Entropy (CE). As a reminder, the CE between $P(y_i)$ and $P(y_i|x_i, \theta)$ in the discrete case is equal to:


\begin{equation}
    CE(P(y_i), P(y_i|x_i, \theta)) = - \frac{1}{N}\sum^N_{i=1} P(y_i) \log P(y_i | x_i, \theta)
\end{equation}

with $P(y_i)=1$ for the labels $y_i \in \mathcal{D}$ whence:  

\begin{align}
    \theta^{MLE}  &= \argmax_\theta -CE(P(y_i), P(y_i|x_i, \theta)) \\&= \argmin_\theta CE(P(y_i), P(y_i|x_i, \theta))
\end{align}

The \textit{Greybox XAI framework} is compositional. Hence, let us decompose $f$ into 2 sub-models $g$ and $h$ such that $f = h \circ g$. 
Let us write $g$ a first model that extracts an \textit{Explainable Latent Space} $\mathcal{Z}$ from the observations $\{x_i\}^N_{i=1}$ and $h$ a second model that maps this Latent Space $\mathcal{Z}$ to labels $\{y_i\}^N_{i=1}$. 
Hence we have: 



\begin{align} \label{Composition}
    & f : \mathcal{X} \rightarrow \mathcal{Y}, f(x_i) =  h(g(x_i, \theta_g), \theta_h) = y_i \\&
     g : \mathcal{X} \rightarrow \mathcal{Z}, g(x_i, \theta_g) = z_i \\&
     h : \mathcal{Z} \rightarrow \mathcal{Y}, h(z_i, \theta_h) = y_i
\end{align}

with $\theta_g$ and $\theta_h$ representing the set of trainable weights of $g$ and $h$. 

Our goal is to map the internal representation of the first model to an \textit{Explainable Latent Space} that will be explainable. In addition, since the prediction of $h$ rely on this \textit{Explainable Latent Space}, we can \adri{ensure} the prediction to be explainable.

This requires not only to have couples $\mathcal{D}_{couple} = \{(x_i, y_i)\}^N_{i=1}$, directly linking images and classes, but rather triples $\mathcal{D}_{triplet} = \{(x_i, z_i, y_i)\}^N_{i=1}$ with $\{z_i\}^N_{i=1} \in \mathcal{Z}^N$ with $\mathcal{Z}$ being an intermediate \textit{Explainable Latent Space} serving as a bridge between $\mathcal{X}$ and $\mathcal{Y}$. Moreover, it is necessary that each element belonging to $\mathcal{Z}$ is a concept that can be expressed in natural language, and thus that this set $\mathcal{Z}$ represents nameable features. This is necessary in order to obtain objective explanations.


The architecture we propose is therefore a compositional model consisting of two elements. First, an opaque DNN called \textit{Latent Space Predictor} denoted $g$ capable of predicting an \textit{Explainable Latent Space} $\mathcal{Z}$. Second, a transparent model called \textit{Transparent Classifier} denoted $h$ able of moving from this \textit{Explainable Latent Space} $\mathcal{Z}$ to a final prediction $\mathcal{Y}$. It results in a framework able, for any image ${x_i}$, to predict which label $y_i$ it corresponds to. This prediction is justified by an objective and intrinsic explanation based on $z_i$ and on the transparent model's simulatability, due to the composition of both models via $\mathcal{Z}$ being the output of $g$ and the input of $h$. It is worth noting that while this framework allows explaining the final prediction $y_i$, based on $z_i$ that acts as the rationale, is not able to explain why $z_i$ was predicted. The \textit{Greybox XAI framework} is therefore a transparent classifier that uses input features from an opaque detector. 



\begin{figure}[h!]
\centering
\includegraphics[scale=0.285]{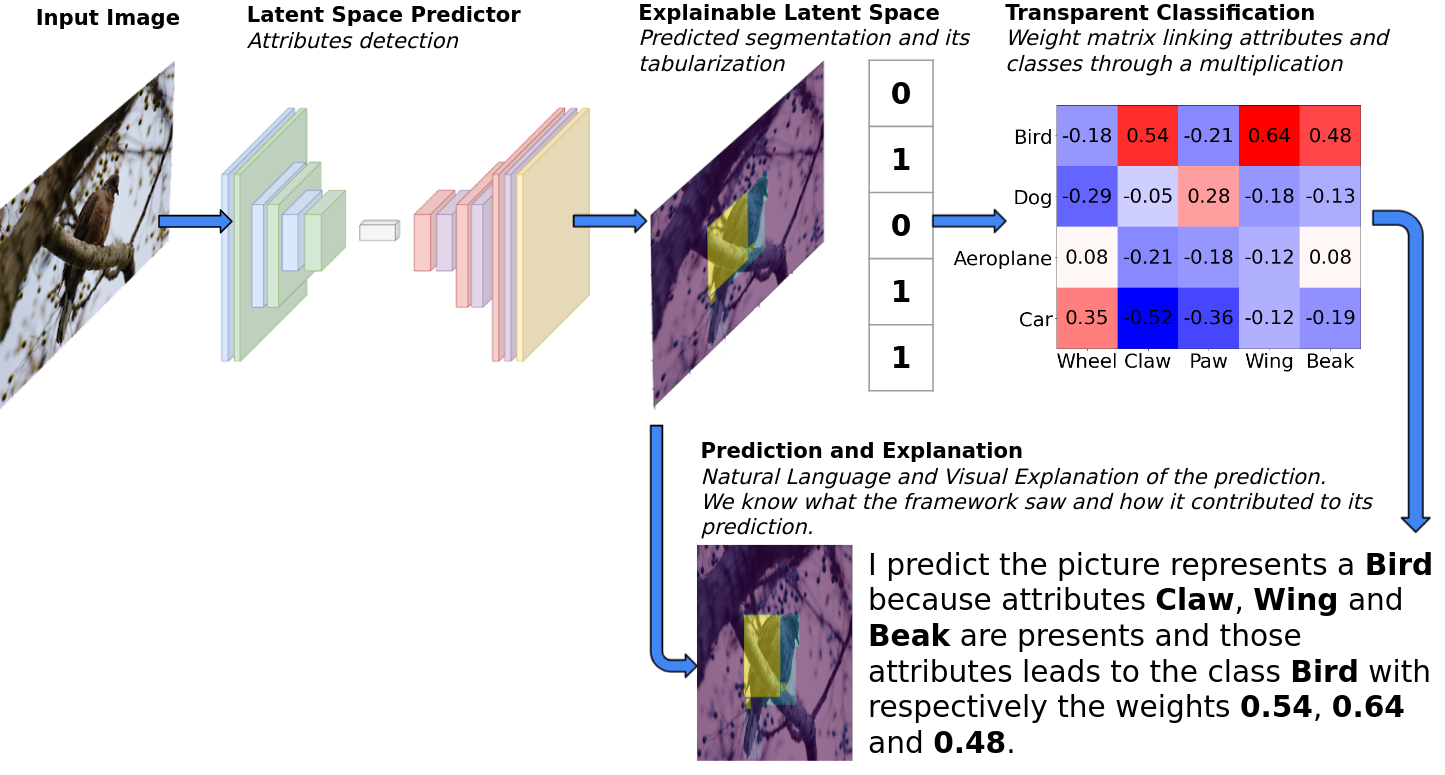} 
\caption{Example of use of the Greybox XAI framework for the task of image classification. The framework produces a prediction of what class of object is present on the image while generating a natural language explanation based on the weight matrix of the transparent classification. It also outputs a visual explanation based on a predicted semantic segmentation image representing different object parts of the predicted class.}
\label{fig:framework}
\end{figure}

Fig. \ref{fig:framework} shows the Greybox XAI framework used for an image classification task. Here, the dataset $\mathcal{D}_{triplet} = \{(x_i, z_i, y_i)\}^N_{i=1}$ consists of triples from a set $\mathcal{X}$ of RGB images, a set $\mathcal{Z}$ of semantic segmentation masks and a set $\mathcal{Y}$ of labels. From $\mathcal{Z}$ we extract a second subset $\mathcal{Z}_{att}$, which we will call the set of attributes. This set contains a list of all attributes, i.e. all different segmentation masks. These are all the \textit{part-of} objects that can be detected by the \textit{Latent Space Predictor}. 


Here the \textit{Latent Space Predictor} that constitutes the model $g$ is an Encoder-Decoder model. Its role is to predict from an image $x_i$ a segmentation map $z_i$. From this segmentation map is extracted the vector $z_{att,i}$ which constitutes a list of all attributes present on the segmentation map $z_i$. We call this operation a vectorization. The couple $\{z_i, z_{att,i}\}$ is the \textit{Explainable Latent Space}. A logistic regression model here acts as a \textit{Transparent Classifier} $h$. \adri{A Naive Bayes Classifier (NB)} can also work, but the experiments conducted in the Section \ref{Experiments} gave better results with logistic regression. We then use the inherent transparent nature of $h$ (developed in Section \ref{Transparent Classifier}) and the \textit{Explainable Latent Space} $\{z_i, z_{att,i}\}$ to explain prediction $y_i$. 





\subsection{Training Process of the Latent Space Predictor and the Transparent Classifier} \label{TrainingProcess}

The \textit{Latent Space Predictor} $g$ and the \textit{
Transparent classifier} $h$ constituting our framework are trained separately. Here are the steps constituting our training: 


\begin{itemize}
\item Manually extract the subset $\mathcal{Z}_{att}$ from $\mathcal{Z}$ in order to have the attributes in the form of a segmentation mask and in the form of a vector. 
\item Train the \textit{Transparent Classifier} to predict $\mathcal{Y}$ using $\mathcal{Z}_{att}$. This model is predicting a class based on an attribute vector.
\item Train the \textit{Latent Space Predictor} to predict latent space $\mathcal{Z}$ using $\mathcal{X}$. This model is predicting a segmentation map based on an RGB image.
\end{itemize}

In Section \ref{Subset Extraction} we explain how we extract the subset $\mathcal{Z}_{att}$. Section \ref{Transparent Classifier} show how we train the \textit{Transparent Classifier} and in Section \ref{LSP} we explain how we train the \textit{Latent Space Predictor}.

\subsubsection{Subset Extraction for Knowledge Base Construction} \label{Subset Extraction}


The goal of the dataset subset extraction task is to obtain a Knowledge Base from a Dataset $\mathcal{D}_{triplet} = \{(x_i, z_i, y_i)\}^N_{i=1}$ with a set $\mathcal{X}$ of RGB images, a set $\mathcal{Z}$ of semantic segmentation masks and a set of (final, whole object) labels $\mathcal{Y}$. We call a Knowledge Base a common repository of semantic annotations to facilitate a fast and efficient search in the given set of resources \cite{Kremen}. 
 



In Description Logics \cite{Baader03basic}, a terminology box \textit{TBox}, or schema extracted from the dataset, together with the class instanciations (in our case, that compose the datasets seen by the network), i.e., the assertional \textit{ABox}, form the $\mathcal{KB} = <TBox, ABox>$. This particular KB will serve as database, in form of a knowledge graph (KG) that interlinks individuals (i.e. instances) of different classes through roles or relationships with possible semantic restrictions. In order to create and populate a Knowledge Base it is necessary that some terminology axioms $TBox$ and assertion axioms $ABox$ can be extracted from some of the training labels, from which a set of Resource Description Framework (RDF) triples of \textit{subject}, \textit{predicate} and \textit{object} ${(s,p,o)}$ can be extracted. This is the case, for example, when the labels are in a text or caption form\footnote{This is also possible for plain \textit{named} class labels (e.g., through using external lexicons or semantic parsers such as \textit{DBPedia} \cite{Auer07} or \textit{Wordnet} \cite{Miller90}.}. The \textit{Automatic Knowledge Base Construction (AKBC) from a dataset} problem consists of finding a process $t: Y_{\mathcal{KB}} \rightarrow \{(s,p,o)\}$ able to \textit{triplify} each data point following the RDF language\footnote{W3C standard model for data exchange: Resource Description Framework \url{w3.org/RDF/}}. Concrete examples of AKBC include learning a KB through semantic parsing, using Markov Logic \cite{Kiddon12} or relational probabilistic models \cite{Balasubramanian12}. The triplification process consists of extracting two entities or concepts, subject (\textit{s}) and object (\textit{o}), and a predicate (\textit{p}) that connects them. This predicate expresses a relationship, i.e., a data property or an object property \cite{Hitzler09} among the subject and the object. The AKBC problem in the \textit{Greybox XAI} framework consists of finding a process that automatically constructs a KB composed of hierarchical (\textit{isA}), \textit{partOf}, and attribute (\textit{hasAttribute}) relations. Analogically, the \textit{ABox} in $\mathcal{KB}$ must be uniquely composed by description logics assertional axioms solely extracted from applying the triplification process on datapoint. 

We flatten each semantic segmentation image $z_i \in \mathcal{Z}$ to obtain a single dimension vector and we return $z_{att,i}$, a sorted list of every unique element contained in $z_i$ in order to obtain a set $\mathcal{Z}_{att}$ of each element present on each semantic segmentation image. By using OWL\footnote{Web Ontology Language (OWL) \cite{antoniou2004web}} entities and relationships we obtain the following type of KB, linking any element $\{z_{att,i, j}\}^{N,K}_{i,j} \in \mathcal{Z}_{att}^{N,K}$ with the corresponding $x_i \in \mathcal{X}^N$ and $y_i \in \mathcal{Y}^N$, with $N$ the number of triple elements of dataset $\mathcal{D}_{triplet}$ and $K$ the number of different attributes in $\mathcal{Z}$:


\begin{table}[h]
\begin{center}
\begin{tabular}{ll}
\textbf{KB}  &\textbf{RDF \textit{(s,p,o)} triple examples} \\
\hline \\
\textbf{TBox} & \texttt{($z_{att,1,1}$, isPartOf, $y_1$)}  \\
 & \texttt{($z_{att,1,2}$, isPartOf, $y_1$)}\\
 & ...\\
 & \texttt{($z_{att,1,k}$, isPartOf, $y_1$)}\\
 & \texttt{($z_{att,2,1}$, isPartOf, $y_2$)}\\
 & ...\\
 & \texttt{($z_{att,n,k}$, isPartOf, $y_n$)}\\
\textbf{ABox} & \texttt{($x_1$, hasLabel, $y_1$)}\\
 & \texttt{($x_1$, hasAttributes, $z_{att,1,1...k}$)}\\
\end{tabular}
\end{center}
\caption{Examples of Resource Description Framework (RDF) triples extracted from a Dataset $\mathcal{D} = (\mathcal{X}, \mathcal{Z}, \mathcal{Y})$ with $n$ the number of samples and $k$ the number of different attributes. These triples are contained in a KB terminological (TBox) and assertional (ABox) components.\label{tab:KB-Dataset}}
\end{table}

This Knowledge Base stands for the \textit{Explainable Latent Space} which will be used to produce the rationale of an explanation. 

Our framework is composed of 2 sequential and connected models that do not perform with the same representation of information: while it is possible for the DNN to work with images, it is necessary for the transparent model to work with a much more reduced and compact representation of the data. It is not possible for a model such as a logistic regression to take images as input because it would lose its transparency due to the exploding number of variables and parameters. A logistic regression needs to have human readable predictors and interactions among them kept to a minimum \cite{arrieta2020explainable}. It can be the case on a compact 2D representation of the semantic segmentation image but not on the image itself because it has too many variables. Therefore, $\mathcal{Z}_{att}$ will be used to train the \textit{Transparent Classifier}. 

\subsubsection{Training of the Transparent Classifier} \label{Transparent Classifier}


We use a logistic regression to statistically fit the attributes $\mathcal{Z}_{att}$ and classes $\mathcal{Y}$ present in the database. The goal is to see if we have fairly discriminating attributes. \adri{The purpose of this test is to check if it is possible to train a transparent classifier based on attributes.  If there were an excessive difference between the performance of a regression predicting classes based on attributes and the performance of a DNN predicting classes based on images, it would mean that the presence/absence of attributes did not allow for an efficient classification of images and therefore that these labels were of poor quality.}
The choice of using a logistic regression is motivated by the fact that this model predicts probabilities easily interpretable \cite{Norton2018}. 

In binary logistic regression the function $h(z_{att,i}, \theta_h)$ used to model the dependence of a regression target $y_{i}\in \{0,1\}$ on features $z_{att,i}$ where $y_{i} \approx h(z_{att,i}, \theta_h)$ can be written as:

\begin{equation}
    h(z_{att,i}, \theta_h) = \frac{1}{1 + \exp(-\theta_h^\intercal z_{att,i})}
\end{equation}

with $\theta_h$ the set of weights trained to minimize the cost function

\begin{equation}
    J(\theta_h) = - \left[\sum_{i=1}^{n}y_{i} \log
    h(z_{att,i}, \theta_h) + (1 - y_{i}) \log (1 - h(z_{att,i}, \theta_h))\right]
\end{equation}

Multinomial logistic regression is a generalization of binary logistic regression to multiclass problems, meaning that the label $y_{i}\in \{1, 2, ..., k\}$ can take $K$ different values depending on the number of classes. A softmax function is used to generalize $h_\theta(z_{att,i})$ from the binary to the multi-class classification problem:

\begin{equation}
     h(z_{att,i}, \theta_h) = 
    \left[
    {
    \begin{array}{c} 
    P(y=1 | z_{att,i} ; \theta_h) \\
    P(y=2 | z_{att,i} ; \theta_h) \\    
    \vdots                    \\
    P(y=k | z_{att,i} ; \theta_h) \\
    \end{array}
    }
    \right]
    =
    \frac
    {1}
    {\sum^k_{j=1} \exp{\theta_{h,j}^\intercal z_{att,i}}}
    \left[
    {
    \begin{array}{c} 
    \exp{\theta_{h,1}^\intercal z_{att,i}} \\
    \exp{\theta_{h,2}^\intercal z_{att,i}} \\    
    \vdots                    \\
    \exp{\theta_{h,k}^\intercal z_{att,i}} \\
    \end{array}
    }
    \right]
\end{equation}

$\theta_{h,1}$, $\theta_{h,2}$, ... $\theta_{h,k} \in \mathcal{R}_n$ are all the parameters of the regression and are represented as a $n$-by-$K$ matrix:

\begin{equation}
    \theta_h
    = 
    \left
    [ 
    {
    \begin{array}{cccc} 
    | & | & | & | \\
    \theta_{h,1} & \theta_{h,2} & \dots & \theta_{h,K} \\    
    | & | & | & | \\
    \end{array}
    }
    \right
    ]
\end{equation}

\begin{equation} \label{cost}
    J(\theta_h) = - \left[\sum_{i=1}^m \sum_{k=1}^k \mathbbm{1}\{y_{i} = k \} \log \frac{\exp{\theta_{h,j}^\intercal z_{att,i}}}{\sum^K_{j=1} \exp{\theta_{h,j}^\intercal z_{att,i}}} 
    \right]
\end{equation}

Equation \ref{cost} is minimized thanks to an iterative optimization algorithm with the gradient: 

\begin{equation}
    \nabla_{\theta_{h,k}} J(\theta_h) = - \sum_{i=1}^m \left[ z_{att,i}\left(\mathbbm{1}\{y_i = k \} - P(y_i=k | z_{att,i} ; \theta_h)\right)\right]
\end{equation}

with 

\begin{equation}
    P(y_i=k | z_{att,i} ; \theta_h) = \frac{\exp{\theta_{h,k}^{\intercal} z_{att_i}}}{\sum^K_{j=1} \exp{\theta_{h,j}^{\intercal} z_{att,i}}}
\end{equation}

Thus, for any $\{z_{att, i}\}^N_{i=1}\in \mathcal{Z}_{att}$ is it possible to find the associated $\{y_i\}^N_{i=1}\in \mathcal{Y}$ by using the following equation:

\begin{equation}
    P(y_i=k | z_{att,i} ; \theta_h) = softmax(\theta_{h,j}^{\intercal} z_{att,i})
\end{equation}

Since the softmax function is monotonic, the ranking of probabilities given by \\ $softmax(\theta_{h,j}^{\intercal} z_{att,i})$ is the same as the one given by $\theta_{h,j}^{\intercal} z_{att,i}$. Therefore, we can approximate that:

\begin{equation}
    h(z_{att,i},\theta_h) \approx  \theta_{h,j}^{\intercal} z_{att,i} \approx \sum_{j=1}^K\theta_{h,j} z_{att,i,j}
\end{equation}

Each parameter $\theta_{h,j}$ provides a quantitative contribution of the corresponding attribute $z_{att,i}$ to predicted class.
This logistic regression is transparent by design because it meets the criteria of algorithmic transparency, decomposability and simulatability from \cite{arrieta2020explainable}: 

\begin{itemize}
    \item Algorithmic Transparency: the user can understand the process followed by the logistic regression to produce any given output from its input data. Just multiply the weight matrix $\theta_h$ by the attribute vector $z_{att,i}$ to obtain the prediction $y_i$.
    \item Decomposability: every part of the logistic regression is understandable by a human without the need for additional tools. The input $z_{att,i}$, the parameters $\theta_h$ and the calculation are interpretable.
    \item Simulatability: the logistic regression has the ability to be totally simulated by a human. It is self-contained enough for a human to think and reason about it as a whole. For this condition to remain true, the complexity of the logistic regression must remain low. Thus, it is necessary that the attribute vector $z_{att,i}$ and the parameters $\theta_h$ are not too big.
\end{itemize}
We can produce explanations following customized templates, for instance:
\begin{example}[Explanation:]
\textit{The model predicts this attribute vector $z_{att,i}$ to belong to class $y_i$, because attributes $\{z_{att,i,1}$, $z_{att,i,...}$, $z_{att,i,k}\}$ are linked to the class $y_i$ with weights $\{\theta_{h,1}$, $\theta_{h,...}$, $\theta_{h_k}\}$ in training dataset $\mathcal{D}_{triples}$.} 
\end{example}

Regarding the definitions expressed in Section \ref{formalisation}, we can say that the explanation function $\Phi(h(z_{att,i})) = \theta_{h}^\intercal z_{att,i}$ of our model $h : \mathcal{Z}_{att} \rightarrow \mathcal{Y}$ is: 

\begin{itemize}
    \item Objective: as the explanation $e_i$ uses symbols ($z_{att,i}$) and relationships ($\theta_{h,i}$) that can be conceptualized by the human user.
    \item Intrinsic: as the explanation $e_i$ only depends of elements ($z_{att,i}$), parameters ($\theta_{h,i}$) and operations present in original model $h$
\end{itemize}

The validity of the explanation will have to be measured when applying the \textit{Greybox XAI} framework to a use-case, as this criterion is dataset-dependant. This notion can be easily measured by comparing the weights $\theta_{h,i}$ and an expert knowledge base.

\subsubsection{Training of the Latent Space Predictor} \label{LSP} 

The \textit{ Latent Space Predictor} must predict a segmentation map $z_i$ from a RGB input image $x_i$ thanks to an Encoder-Decoder architecture. This segmentation map will then be vectorized in an attribute vector $z_{att,i}$ to constitute the \textit{Explainable Latent Space} $\{z_i, z_{att,i}\}$. We choose to use a DeepLabv3+ \cite{chen2018encoderdecoder} as a \textit{Latent Space Predictor} with a ResNet101 as backbone model. The specificity of DeepLabv3+ is to use an atrous convolution, allowing the developer to adjust filter's field-of-view in order to capture multi-scale information, and a depthwise separable convolution. As it is a semantic segmentation task, we use an output stride of 16 for denser feature extraction as it is the best trade-off between speed and accuracy. The objective is to make a pixel-wise prediction over an entire image and the performance is measured in terms of pixel intersection-over-union averaged across the attributes (mIOU).

In order to test the performance of our model, we do not train it with images of semantic segmentation masks but with bounding boxes showing the presence or absence of attributes, in a weekly supervised manner \cite{kervadec2020bounding}. The main objective is to have an \textit{Explainable Latent Space} that accounts for the presence of each attribute as much as possible. It is therefore necessary that each attribute present on the $x_i$ image is segmented but the segmentation map does not need to be very accurate because it is afterwards vectorized. 
Pixels that do not belong to any bounding box are considered as background pixels, while the ones belonging to several bounding boxes 
are accounted as belonging only to the smallest one. We select the smallest rather than the largest in order to not lose the small attributes encompassed by large ones (like eyes in the middle of a face for example). 


The input of the Latent Space predictor is an RGB image $x_i$ while its output is a segmentation map $z_{i}$ of dimension $h*w$ with $h$ and $w$ the dimensions of $x_i$. As the spatial information is not used by the \textit{Transparent Classifier}, we extract from $z_{i}$ an attribute vector $z_{att,i}$ containing the list of each unique value contained in $z_{i}$. A confidence mask is applied in order to keep only the attributes that were predicted with a confidence above a certain threshold. We finally use a one-hot encoding to obtain a vector of 0s and 1s, describing the prediction of presence or absence of each attribute in the input RGB image $x_i$. Because this extraction does not allow backpropagation, the \textit{Latent Space Predictor} is trained by maximising its mIOU. It is therefore not directly trained to predict a good attribute vector $z_{att,i}$ but a good segmentation map $z_{i}$. Note that this segmentation map prediction is opaque, no explanation is given as to why a certain pixel has been predicted as representing a certain attribute.

\subsection{Inference prediction and its explanation rendering through a Natural Language Explanation} \label{Inference}

When the \textit{Latent Space Predictor} $g$ and the \textit{Transparent Classifier} $h$ are trained and provide good results on their own, we freeze their weights 
and compose them to evaluate the function $(h \circ g) : \mathcal{X} \mathcal{Y}$ in order to predict a class $y_i$ from $x_i$. As the \textit{Transparent Classifier} $h$ is transparent by design, we are able to generate a natural language explanation while making the prediction $y_i = h(g(x_i))$. 
As explained earlier we produce an explanation of prediction $\hat{y}$ from 3 elements: 
\begin{itemize}
    \item The intrinsic transparency of $h$, allowing to know what would imply a change of $z_{att,i}$ on the final prediction $\hat{y}$ thanks to the learned weights in $\theta_h$.
    \item $z_{att,i}$ which takes the form of a list of attributes that can be named in natural language.
    \item $z_{i}$ which is a segmentation map, showing the position of attributes on the RGB image $x_i$, thus taking over the ease of understanding of the usual visual explanations. 
\end{itemize}

We define the explanation function $\Phi : \mathcal{Y} \rightarrow \mathcal{E}$ with $ \mathcal{Y}$ the label space and $\mathcal{E}$ the explanation space. The \textit{Transparent Classifier} $h$ and the \textit{Explainable Latent Space} $\{z_{i}, z_{att,i}\}$ make it possible to produce explanation $e$ in natural language.

\begin{equation}
    e_i = \Phi(h(z_{att,i}))
\end{equation}

Therefore, we have the following explanation:

$e_i$ = "Image $x_i$ represents a $y_i$ because attributes $z_{att,i,1}$, $z_{att,i,...}$ and $z_{att,i,m}$ are present, and the classifier $h$ leads those attributes respectively with weights $\theta_{h,1}$, $\theta_{h,...}$ and $\theta_{h,n}$ to class $y_i$." 

In addition, the segmentation map $z_i$ can be displayed as a visual explanation to show the position of attributes $z_{att,i}$.



As a summary, the \textit{Greybox XAI} framework makes a prediction $y_i$ of a random RGB image $x_i \in X$ and produces an explanation $e_i$ by following Algorithm \ref{prediction}:

\begin{algorithm} 
\caption{Greybox XAI framework pseudo-algorithm to produce a natural language explanation of a prediction}
\begin{algorithmic}[1]\label{prediction}
\REQUIRE Input Image $x_i$, 
Latent Space Predictor $g$, Transparent Classifier $h$, Explanation Function $\Phi$ 
\STATE Step 1: Predict Explainable Latent Space
\STATE ${z_i} \leftarrow g(x_i)$
\STATE Step 2: Vectorize Explainable Latent Space to obtain an attribute vector
\FOR{$j \in {z_i}$}
\STATE Append(${z_{att,i}}$, $j$) if $j \notin {z_{att,i}}$
\ENDFOR
\STATE Step 3: Predict Object Class
\STATE $y_i \leftarrow h(z_{att,i})$
\STATE Step 4: Generate a Natural Language Explanation 
\STATE $e_i \leftarrow \Phi(h(z_{att,i}))$
\RETURN Prediction $y_i$, Natural Language Explanation $e_i$, Segmented Image ${z_i}$
\end{algorithmic}
\end{algorithm}


\section{Experimental Study} \label{Experiments}


We illustrate the use of our framework with 2 datasets: MonuMAI, PASCAL-Part. The extensive use case proving the utility of the model is developed on MonuMAI because this dataset has already been used in the state of the art \cite{Diaz-Rodriguez21} to prove the utility of compositional models. The hypothesis tested is to verify that the \textit{Greybox XAI} framework is able to produce accurate and explainable predictions. Our goal is to solve a compositional classification problem and to be able to predict for each image which object is present, justifying this prediction by the \textit{parts-of} object (attributes) of this object present on the image. 

MonuMAI dataset \cite{lamas2020monumai} allows to classify architectural style classification from facade images. The idea here is to be able to classify an image by predicting which type of monument is present in the image based on the distinctive architectural attributes of the different types of monuments. This dataset contains approximately 1500 images labelled with 4 classes (architectural styles) and containing bounding boxes describing the presence of 15 different attributes (visible characteristics of these architectural styles). Each image is labelled with the architectural style of the monument present on the image and bounding boxes inform about the presence and position of the attributes present on the image. We call this dataset $\mathcal{D}_{triple} = \{(x_i, z_i, y_i)\}^N_{i=1}$ with a set $\mathcal{X}$ of RGB images representing architectural monuments, a set $\mathcal{Z}$ of bounding boxes representing architectural attributes and a set $\mathcal{Y}$ of architectural styles.

A Knowledge Base \ref{tab:KB-MonuMAI} is built based on the expert knowledge of the MonuMAI dataset \cite{lamas2020monumai}. 

\begin{table}[htbp!]
\begin{center}
\begin{tabular}{ll}
\textbf{KB}  &\textbf{RDF \textit{(s,p,o)} triple examples} \\
\hline \\
\textbf{TBox} & \texttt{(Ogee Arch, isPartOf, Gothic Monument)}  \\
 & \texttt{(Pointed Arch, isPartOf, Gothic Monument)}\\
 & \texttt{(Trefoil Arch, isPartOf, Gothic Monument)}\\
 & \texttt{(Gothic Pinnacle, isPartOf, Gothic Monument)}\\
 & \texttt{(Flat Arch, isPartOf, Hispanic-Muslim Monument)}\\
 & \texttt{(Lobed Arch, isPartOf, Hispanic-Muslim Monument)}\\
 & \texttt{(Horseshoe Arch, isPartOf, Hispanic-Muslim Monument)}\\
 & \texttt{(Broken Pediment Arch, isPartOf, Baroque Monument)}\\
 & \texttt{(Solomonic Column, isPartOf, Baroque Monument)}\\
 & \texttt{(Rounded Arch, isPartOf, Baroque Monument)}\\
 & \texttt{(Rounded Arch, isPartOf, Renaissance Monument)}\\
 & \texttt{(Porthole Arch, isPartOf, Baroque Monument)}\\
 & \texttt{(Porthole Arch, isPartOf, Renaissance Monument)}\\
 & \texttt{(Lintelled Doorway Arch, isPartOf, Baroque Monument)}\\
 & \texttt{(Lintelled Doorway Arch, isPartOf, Renaissance Monument)}\\
 & \texttt{(Serliana, isPartOf, Renaissance Monument)}\\
 & \texttt{(Segmental Pediment, isPartOf, Renaissance Monument)}\\
 & \texttt{(Triangular Pediment, isPartOf, Renaissance Monument)}\\
\end{tabular}
\end{center}
\caption{Examples of RDF triples extracted from the MonuMAI dataset and contained in a KB terminological (TBox) and assertional (ABox) components} \label{tab:KB-MonuMAI}
\end{table}

We repeat the various steps described in the Section \ref{TrainingProcess} in order to train and use the \textit{Greybox XAI} framework:

\begin{itemize}

    \item A transparent model $h$ is trained to predict an architectural style, using as input a vector encoding the presence and absence of architectural attributes.
    \item A Deep Neural Network $g$ is trained to predict a segmentation map from an RGB image input. Its purpose is to detect the different architectural attributes that constitute the image.
\end{itemize}

\subsection{Logistic Regression as a Transparent Classifier} \label{LogReg}

The purpose of the \textit{Transparent Classifier} is to represent a more accurate and closer version of the dataset than the Knowledge Base \ref{tab:KB-MonuMAI} itself. In fact, the knowledge contained in the knowledge base is very generic (for example, a Hispanic-Muslim monument has a flat arch, an horseshoe arch and a lobed arch). However, while this is true in a general case, not all such monuments have all these attributes and some representations (i.e. images) of these monuments may have some attributes missing. Also, in this particular case of monuments, it is possible that some images have architectural attributes belonging to several architectural styles. It is the result of the progressive evolution of the construction or reconstruction processes. 


In binary logistic regression the function $h(z_{att}, \theta_h)$ used to model the dependence of a regression target $y_{i}\in \{0,1\}$ on features $z_{att,i}$ where $y_{i} \approx h(z_{att,i}, \theta_h)$ can be written as:

Following Section \ref{Transparent Classifier}, a logistic regression $h(z_{att}, \theta_h)$ is trained on the \adri{set of attributes $\mathcal{z}_{att}$} of the dataset to predict classes from the set of labels $\mathcal{Y}$. 
We compare the performance of the logistic regression to a Naive Bayes Classifier. 

\begin{table}[htb]
\centering
\label{t:MeanAccuracy}
\begin{tabular}{lcc}
\noalign{\smallskip} \hline \hline \noalign{\smallskip}
\textbf{Model} & \textbf{Accuracy} \\
\hline
\textbf{Logistic Regression} & \textbf{97.65}\% \\
Naive Bayes Classifier & 94.32\% \\
\noalign{\smallskip} \hline \noalign{\smallskip}
\end{tabular}
\caption{Mean Accuracy of 2 transparent models on MonuMAI dataset. The logistic regression have a better accuracy than the Naive Bayes Classifier. We therefore choose to use it as the \textit{Transparent Classifier} of the \textit{Greybox XAI} framework}
\end{table}

\begin{figure*}[htbp!]
\centering
\includegraphics[scale=0.28]{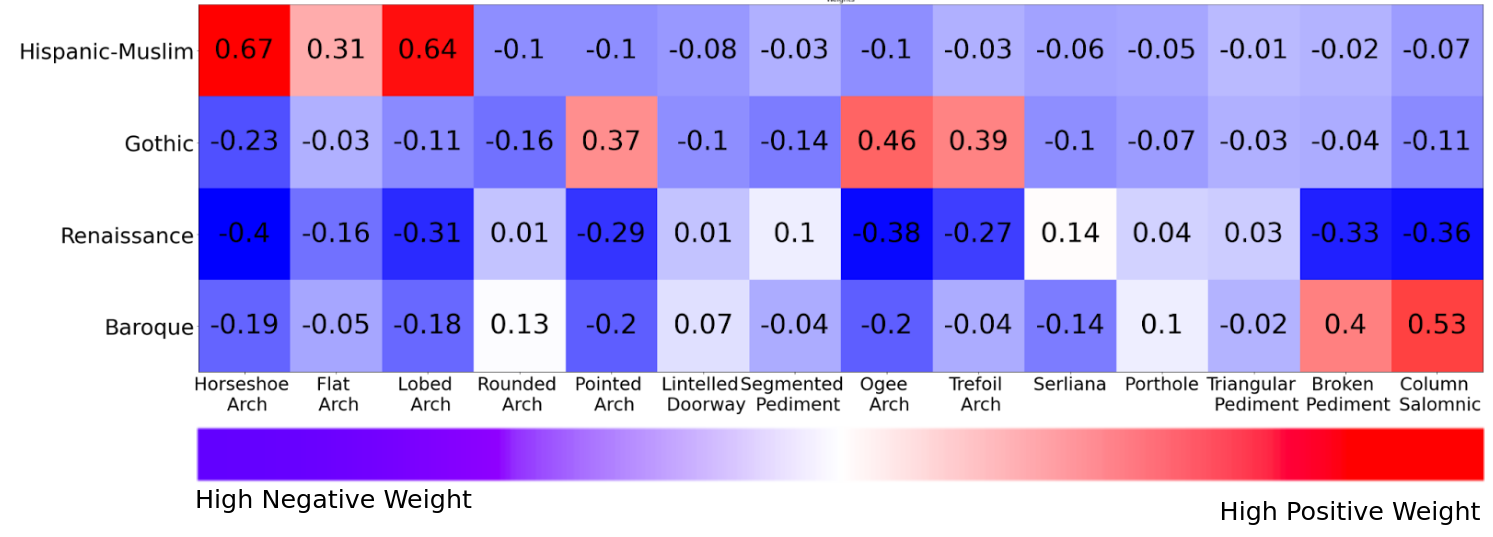} 
\caption{Weights of the logistic regression model fitted to link attributes and classes from the MonuMAI dataset.} 
\label{fig:weight_linreg_monumai}
\end{figure*}

Figure \ref{fig:weight_linreg_monumai} represent the set of trainable weights $\theta_h$ of the logistic regression, linking attributes from ${z_{att}}$ and classes from $\mathcal{Y}$ thanks to the relationship $y_i \approx \theta_h ^\intercal \times z_{att_i}$. These weights provide a statistical link between attributes and classes. 


\begin{figure*}[htbp!]
\centering
\includegraphics[scale=0.3]{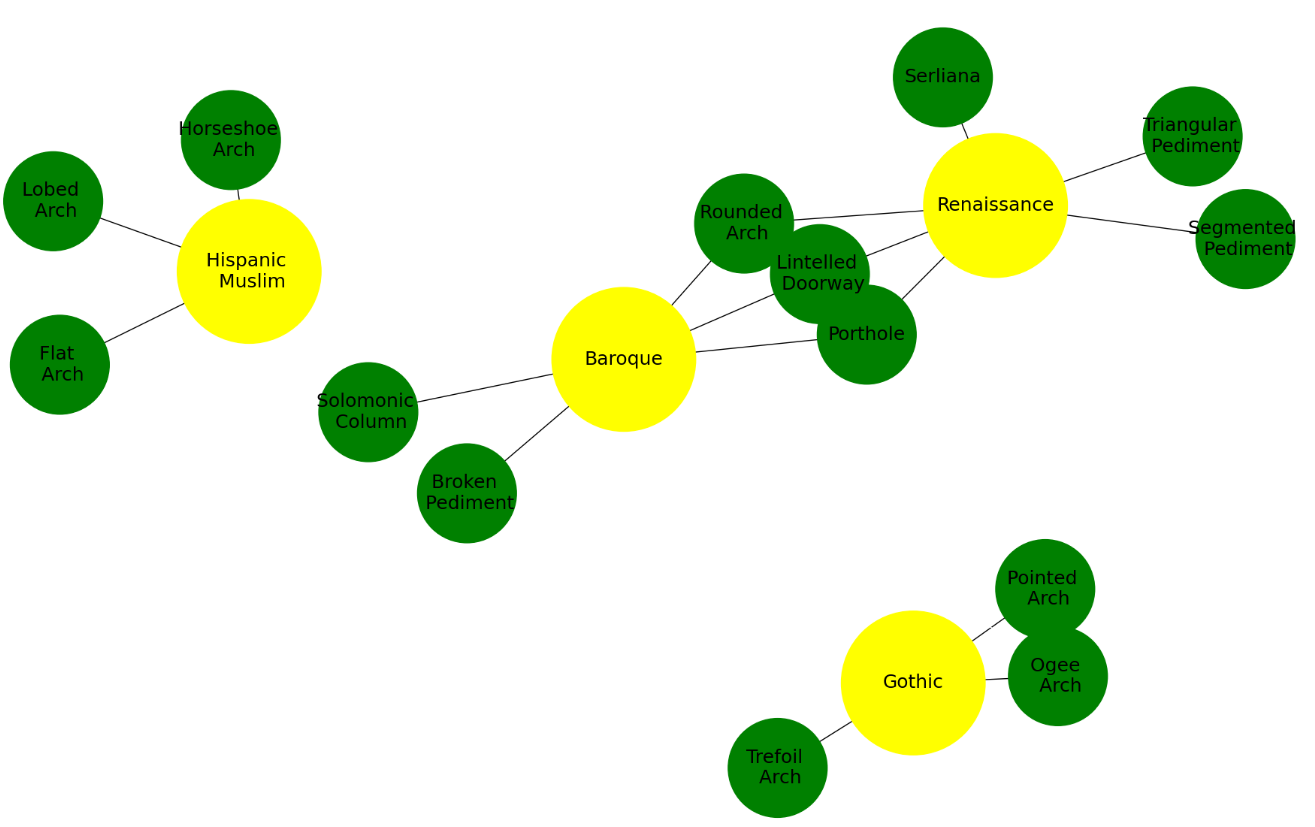} 
\caption{Knowledge Graph representation automatically extracted from the Logistic Regression weights on the MonumAI dataset. Green nodes represent attributes while yellow nodes represent classes. Distances between nodes represent weights linking attributes and classes in the $\theta$ matrix of the fitted logistic regression model: the closer 2 nodes are, the larger the weight linking them. An edge is set to black if the associated weight is superior to zero and transparent otherwise.}
\label{fig:KG_Monumai}
\end{figure*}

From the set of trainable weights $\theta_h$ of the \textit{Transparent Classifier} we extract a Knowledge Graph (KG) (Figure \ref{fig:KG_Monumai}) representing the link between attributes and classes. It is a visual representation of the weights from Figure \ref{fig:weight_linreg_monumai}. \adri{If a weight is superior to 0, an edge is drawn between the 2 concerned nodes.} Representing knowledge this way has a simple explanatory interest: when an attribute is detected, it is straightforward to see which classes are linked to this attribute. We see for example that the attributes \textit{Trefoil Arch}, \textit{Pointed Arch} and \textit{Ogee Arch} are only linked to the architectural style \textit{Gothic}. Therefore, if those attributes are present in the vector $z_{att_i}$, the class \textit{Gothic} will be predicted by the \textit{Transparent Classifier}.

\subsection{Deeplabv3+ as a Latent Space Predictor} \label{Deeplab}

Logistic regression cannot remain transparent when taking images as input because the number of parameters and variables would be far too large. Moreover, these variables would be pixels rather than symbols. To overcome this and to make the logistic regression take as input a vector of attributes, we train an Encoder-Decoder on the images on a semantic segmentation task. We chose to use a DeepLabv3+ \cite{chen2018encoderdecoder} as it gives the best performance on this dataset.


On this dataset we do not have a semantic segmentation image representing the ground truth but only bounding boxes around each attribute. We use these bounding boxes to predict a segmentation map thanks to a cross-entropy loss. As the model was trained with images annotated with bounding boxes instead of semantic segmentation images, the segmented attributes have a square shape. It is not a problem as unlike most semantic segmentation models, what we are interested in here is not the mIoU or the accuracy of the prediction at the pixel level but rather how well the attribute vector $z_{att,i}$ is predicted. Since the \textit{Transparent Classifier} takes as input a one-hot encoded vector of attributes, the mIoU and the average precision of the Deeplabv3+ have no influence on the classification result\adri{: what Greybox model focuses on is on detecting at least once each attribute, not on detecting all pixels of each occurrence of each attribute. Whether we detect one pixel of a given attribute or 3000 pixels of an attribute is the same because the segmentation map is put in the form of a binary vector of attribute presence in order to be used by the \textit{Transparent Classifier}.}

\begin{figure*}[htbp!]
\centering
\includegraphics[scale=0.80]{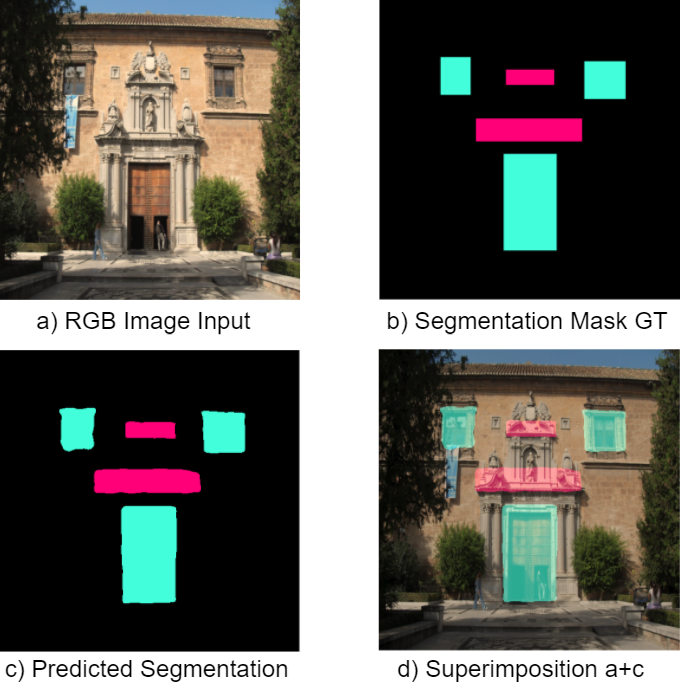} 
\caption{Example of results obtained using semantic segmentation. Image a) represents the RGB input data and Image b) represents the Ground Truth of semantic segmentation masks. Image c) represents the result of this segmentation by the \textit{Latent Space Predictor} and image d) is an overlay of the prediction on the input image, in order to see where the detected attributes are. The black pixels represent the background, the cyan and pink pixels represent two architectural attributes.}
\label{fig:Deeplab}
\end{figure*}

To generate this attribute vector, we compare the list of sorted unique elements of the predicted semantic segmentation image and the list of attributes present in the image. 
Figure \ref{fig:Deeplab} represents on the left an RGB image used as input to the Deeplabv3+ and on the right the predicted semantic segmentation image. The attribute vector associated with this semantic segmentation image would be, for example, $[1, 0, 1, 1, 0, ..., 0]$ representing the 3 attributes (in dark cyan, pink and blue) present in the image. It is this vector that is subsequently used by the pre-trained \textit{Transparent Classifier} to predict the class of the image. 


\subsection{Performance of the Greybox XAI framework: Accuracy and Explainability} \label{Performance}

We judge the performance of our framework according to 2 notions: its accuracy during an image classification task and the explainability of its prediction during this same classification task.

\subsubsection{Accuracy}

We evaluate our model on the image classification task and compare it to several baselines. Our model is the \textit{Greybox XAI Framework}, composed by a Deeplabv3+ and a logistic regression. 
\adrien{The first baseline is a Data-efficient image Transformers (DeiT) \cite{Touvron2020}, a convolution-free transformer built upon the Vision Transformer (ViT) architecture \cite{dosovitskiy2020vit} and achieving state-of-the-art results over baseline datasets. In order to obtain the best performing comparison baseline, we used the largest possible DeiT architecture (similar to that of VIT-B), as well as the highest possible image resolution (384x384).} The second baseline is a ResNet101 classifier, in order to have a convolution-based baseline. We then compare to the EXPLANet \cite{Diaz-Rodriguez21} model which is the state of the art of compositional XAI models on this database. We also build a new baseline by modifying the DeepLabv3+ to make it a classifier. This Deeplab Classifier is a usual Deeplabv3+ architecture, with a ResNet101 as backbone, with skip connections and atrous convolutions, but instead of predicting the class of each pixel as in a semantic segmentation we use it in a classification role by modifying the last layer. Instead of evaluating it by the MiOU, it is now judged by the accuracy of the global class. By adding an AveragePooling and a Softmax after the decoder, we obtain an end-to-end classification model. Since this is a classification task, we compare ourselves in terms of accuracy. We also add the results on the PASCAL-Part dataset, which contains more attributes and more classes. As our model is transparent by design, we compare ourselves separately to models considered as opaque (Transformers like DeiT and CNNs such as ResNet or Encoder-Decoder such as Deeplabv3+) and explainable models (such as EXPLANet). 

The results of Greybox XAI\footnote{\url{https://github.com/AdrienBennetot/Greybox-}}
classification model based on Deeplabv3+ semantic segmentation backbone and a logistic regression, together with these baseline classification networks are shown in Table \ref{tab:results} for MonuMAI and PASCAL-Part dataset. The results show that the Greybox XAI classifier achieves accuracy slightly inferior to the opaque models, being outperformed by approximately 2.5\% each time. However, its accuracy is far superior to the explainable baseline, which is the EXPLANet model. We can therefore consider that there is a slight loss in accuracy compared to the opaque models but a gain compared to the compositional models.

\begin{table}[]
\centering
\begin{tabular}{lll}
\textbf{Dataset}    & \textbf{Model}                                     & \textbf{Accuracy (\%)}                     \\ \hline
\multicolumn{3}{l}{\textbf{Comparison with Opaque Models}}                                                       \\ \hline
\textbf{MonuMAI}    & Greybox XAI (our)                                       & 94.04                                 \\
                    & Deeplabv3+ Classifier                            & 96.02                     \\
                    & \textbf{DeiT-B}                            & \textbf{96.48}                       \\
                    & ResNet101                                            & 95.69                                 \\
                    & MonuNet Classifier                                 & 83.11                                 \\ \hline
\textbf{PASCAL-Part} & Greybox XAI (our)                                       & 88.30                                 \\
                    & Deeplabv3+ Classifier                            & 90.18                      \\
                    & \textbf{DeiT-B}                            & \textbf{90.85}                       \\
                    &  ResNet101                   & 90.12
                    
                                \\ \hline
\multicolumn{3}{l}{\textbf{Comparison with Explainable Models}}                                                  \\ \hline
\textbf{MonuMAI}    & \textbf{Greybox XAI (our)       }                                  & \textbf{94.04 }                                \\
                    & KG Deterministic Classifier & 54.79       \\
                    & EXPLANet                                             & 90.40                      \\ \hline
\textbf{PASCAL-Part} &    \textbf{Greybox XAI (our)}                                        & \textbf{88.30}                                  \\
                    & EXPLANet                     & 82.4
\end{tabular}
\caption{Explainable compositional vs opaque direct classification: Results of the Greybox XAI model (using semantic segmentation Deeplabv3+ and a logistic regression) on MonuMAI and PASCAL-Part datasets, and comparison with embedded version of the baseline model MonuNet, a vanilla classifier baseline with ResNet101, a transformer model with DeiT, an expert KG-based deterministic (non-trained) classifier, the compositional model EXPLANet and a classifier derived from Deeplabv3+ 
}
\label{tab:results}
\end{table}

The loss in accuracy compared to opaque models should be counterbalanced by a benefit in explainability because the \textit{Greybox XAI} is transparent by design when used for a classification task (see Section \ref{Transparent Classifier}) and produces "good" explanations (\textit{objective} and \textit{intrinsic}). 

\subsubsection{Explainability}

\begin{figure}[htbp!]
\centering
\includegraphics[scale=0.4]{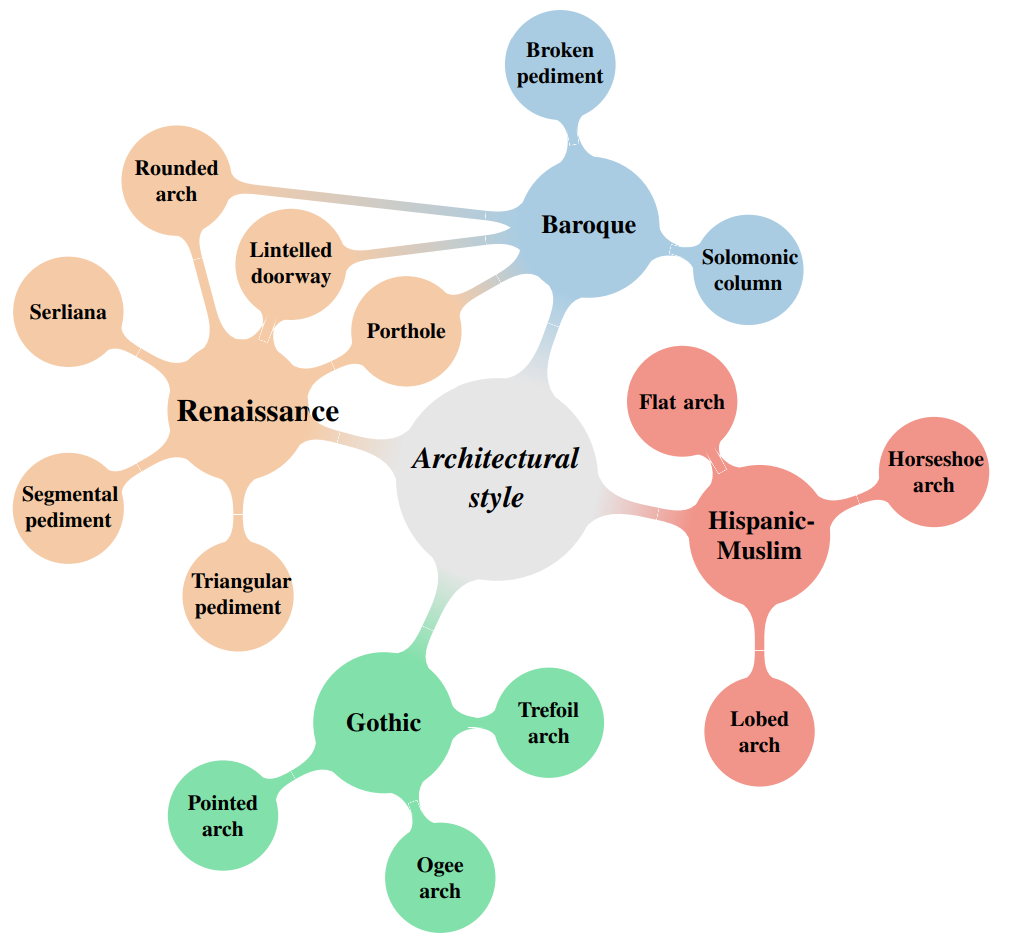} 
\caption{Simplified MonuMAI knowledge graph constructed based on art historians expert knowledge \cite{Diaz-Rodriguez21, lamas2020monumai}. It links attributes and classes in a graphical representation and shows, for example, that a Hispanic-Muslim monument is composed by Flat Arches, Horseshoe Arches and Lobed Arches. This graph representing the expert knowledge of the dataset is similar to the one extracted by the \textit{Transparent Classifier}.}
\label{fig:kb_monumai}
\end{figure}

In order to verify whether the explanations generated by the \textit{Greybox XAI} are \textit{valid} we compare the Knowledge Graph extracted directly from the \textit{Transparent Classifier} weights in Figure \ref{fig:KG_Monumai} and the expert Knowledge Graph Figure \ref{fig:kb_monumai}. This figure is taken from \cite{Diaz-Rodriguez21} and represents MonuMAI Knowledge Graph constructed based on art historians expert knowledge \cite{lamas2020monumai}. The KG is extracted from the \textit{Transparent Classifier} by creating a node for every \textit{part-of} and \textit{objects} of the weights matrix and by drawing an edge between nodes for every weights superior to 0.
The Graph Edit Distance (GED) \cite{sanfeliu1983distance} between the two KG is equal to zero, meaning that the explanation of the \textit{Transparent Classifier} is the same as the ones art historians experts would have produced. Therefore, we can consider that these explanations are \adri{globally} \textit{valid} \adri{if the semantic segmentation is correct}.  

We also verify if our framework is a \textit{self-explaining prediction model} according to the definition of \cite{Alvarez-Melis18}, i.e. if it has the form:

\begin{equation}\label{eq:alvarez}
    f(x) = g(\theta(x)_1 h(x)_1, ..., \theta(x)_k h(x)_k)
\end{equation}

where:

\begin{itemize}
    \item $g$ is monotone and completely additively separable
    \item  For every $z_i:= \theta_i(x)h_i(x)$, $g$ satisfies $\frac{\partial g}{\partial z_i} \geq$ 0
    \item  $\theta$ is locally difference bounded by $h$
    \item  $h_i(x)$ is an interpretable representation of $x$
    \item  $k$ is small
\end{itemize}

Taking Equation \ref{Composition}, \textit{Greybox XAI} framework can be written:

\begin{equation}\label{eq:alvarez}
    f(x) = (h \circ g)(x) =  h(g(x_i, \theta_g), \theta_h) \approx \theta_{h,1} z_{att,1},...,\theta_{h,n} z_{att,n}
\end{equation}

where:
\begin{itemize}
    \item The \textit{Transparent Classifier} $h$ is monotone and completely additive separable as it can be approximated with the multiplication between the weight matrix $\theta_h$ and the features.
    \item Partial derivative of $h$ with respect to $\theta_{h,i} z_{att,i}$ is positive.
    \item $\theta_h$ is locally difference-bounded by $z_{att,i}$.
    \item $z_{att,i}$ is an interpretable representation of $x$ as $z_{att,i}$ are nameable features.
    \item $n=$ is small as interactions of the logistic regression are kept to a minimum to respect the definition of simulatability.
\end{itemize}

\begin{figure*}[htbp!]
\centering
\includegraphics[scale=0.45]{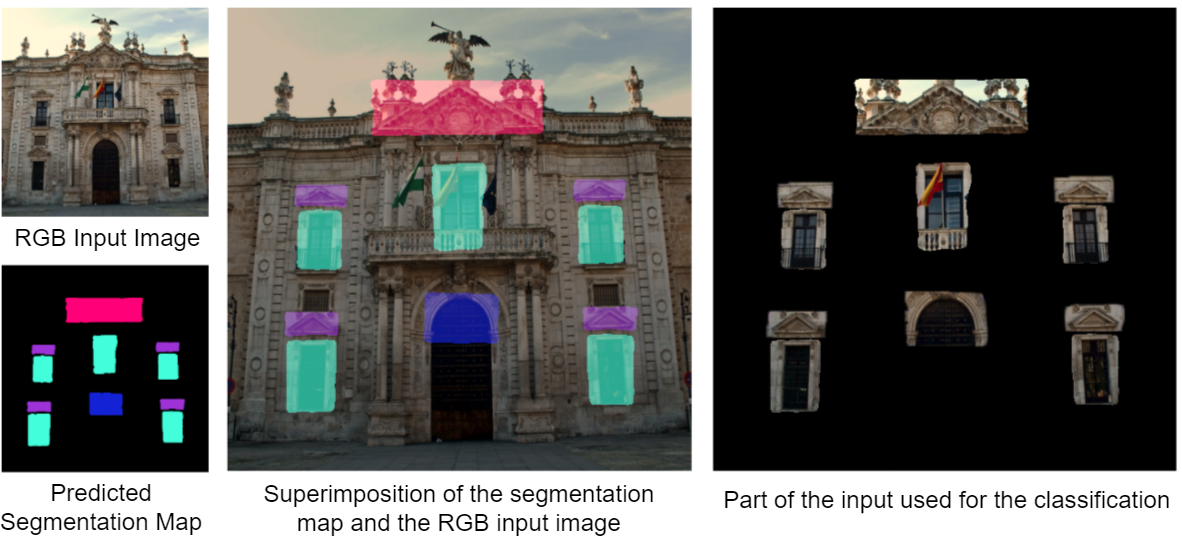} 
\caption{Visual explanation of the Greybox XAI model. The image in the upper left corner is the RGB input that the model must classify. In the bottom left corner is the semantic segmentation image predicted by the model, showing in \adri{black} the pixels classified as part of the background. Cyan and black pixels represents two attributes. In the middle the superposition of the two images on the left, removing \adri{black} pixels from the background to keep only the elements that will be used in the logistic regression. Finally, the image on the right is the same sample image but replacing the attributes by their RGB value and hiding the background pixels, which are not used during the classification by the Transparent Classifier \adri{as it only uses as input an attribute vector.}} 
\label{fig:Deeplab segmentation}
\end{figure*}

From these different elements, we can conclude that the Greybox XAI model is transparent and produces "good" explanations when used for the task of image classification. Moreover, it is possible to accompany this textual explanation by a visualization, showing the semantic segmentation image masks used to determine the attribute vector $\mathcal{z}_{att}$ employed by the \textit{Transparent Classifier} to perform the classification. Figure \ref{fig:Deeplab segmentation} shows an example of the visualization that can be produced. This image has been classified as \textit{Gothic} based on the different attributes detected by the Latent Space Predictor. The predicted semantic segmentation map can be overlaid on top of the RGB input to visualize where these features are located. The image on the right shows all pixels that were used during classification. These are obtained by removing every pixel segmented as being a background pixel.

Figure \ref{fig:Deeplab segmentation} illustrates the ideal case where the semantic segmentation performed is perfect and the ensuing classification is also perfect. However, there are some cases where the prediction of the segmentation map is either flawed or false, which is sometimes inconsequential but that will sometimes distort the prediction. Below we present the different cases observed and associated example.

\begin{example}[Incomplete segmentation, correct prediction:]
    
Some occurrences of an attribute are not detected. As vectorization does not take into account the number of occurrences of each attribute, this has no impact on the prediction or the explanation. However, the predicted segmentation map is far from the real segmentation map, which gives an incomplete visual explanation. \adri{See Figure \ref{fig:example_1}. This is the kind of example that ultimately has no impact on the result: the prediction remains \adri{correct} and the explanation is valid and complete because it addresses the 2 main elements detected. The fact that some occurrences of one of the attributes are forgotten is of little importance for our classification task. However, for a detection task where detecting each and every instance of an object type is critical, it could have been very problematic. A future improvement could be using an Optimized Loss Function for Object detection \cite{Jiang_2021} when training the \textit{Latent Space Predictor} in order to improve the detection.}

\begin{figure*}[htbp!]
\centering
\includegraphics[scale=0.40]{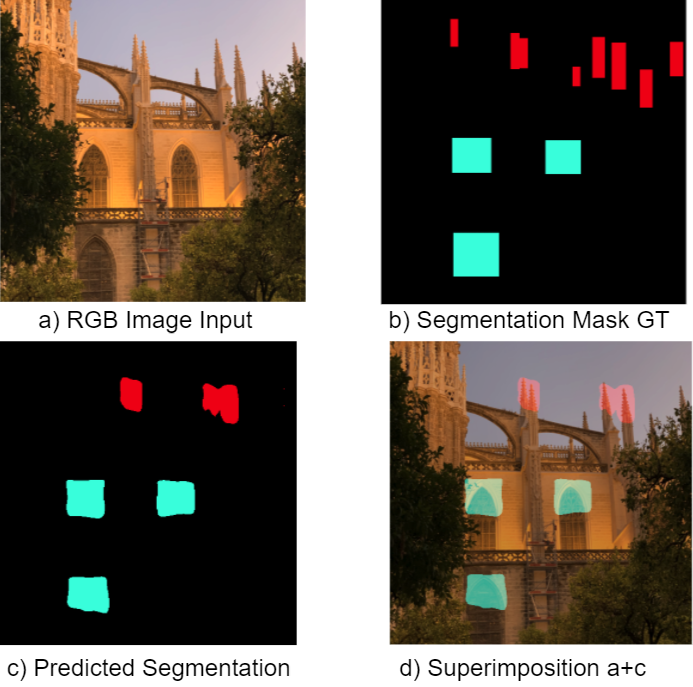} 
\caption{\adri{\textbf{Incomplete segmentation, correct prediction}}. Example of results obtained using semantic segmentation. Image a) represents the RGB input data and Image b) represents the semantic segmentation masks of the Ground Truth, which the Latent Space Predictor must predict. Image c) represents the result of this segmentation and image d) is an overlay of the prediction on the input image, in order to see where the detected attributes are. Black pixels represent the background, cyan and red pixels represent two architectural attributes. The red attribute is missing 5 times compared to the GT but it has no impact on the prediction nor the explanation.}
\label{fig:example_1}
\end{figure*}
\end{example}

\begin{example}[Wrong segmentation, correct prediction:]

\adri{Some attributes are missing but the classification does not change as the detected attributes are already discriminative enough for the \textit{Transparent Classifier}. See Figure \ref{fig:example_2}. Here it is complicated to judge the validity of the explanation without being an expert: is the explanation complete enough by talking about the cyan and red attributes or was the blue attribute indispensable for the classification of this monument? Looking at the \textit{Transparent Classifier} weight matrix, we can see how important the blue attribute should have been in the prediction. A higher quality segmentation would correct this case of model failure: if the blue attribute had been detected, the image would have been predicted correctly.} 

\begin{figure*}[htbp!]
\centering
\includegraphics[scale=0.40]{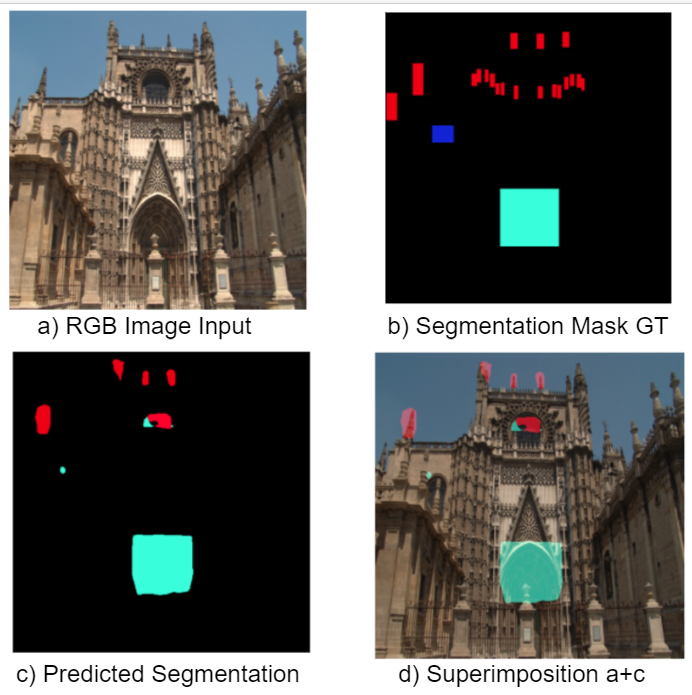} 
\caption{\adri{\textbf{Wrong segmentation, correct prediction}}. Example of results obtained using semantic segmentation. Image a) represents the RGB input data and Image b) represents the semantic segmentation masks of the Ground Truth, which the Latent Space Predictor must predict. Image c) represents the result of this segmentation and image d) is an overlay of the prediction on the input image, in order to see where the detected attributes are. The black pixels represent the background, the colored pixels represent architectural attributes. The red attribute is missing several times and the blue one is totally absent compared to the GT but it has no impact on the prediction, as the presence of the red and cyan elements are enough to consider that this monument is Gothic.}
\label{fig:example_2}
\end{figure*}
    \end{example}

\begin{example}[Wrong segmentation, wrong prediction:]

\adri{Some attributes are detected while they are not existing in the ground truth. It makes the classification totally wrong. See Figure \ref{fig:example_3}. This error is more problematic than the previous one because it misleads the user by giving an invalid explanation. The model is not \textit{Right for the Right Reason}. This is primarily a semantic segmentation problem: the model should not have seen the yellow attribute and the cyan attribute. The model could thus be improved, perhaps with the use of a weighted FocalLoss\cite{Qin_2018} to penalize more strongly a wrong attribute detection.}

\begin{figure*}[htbp!]
\centering
\includegraphics[scale=0.40]{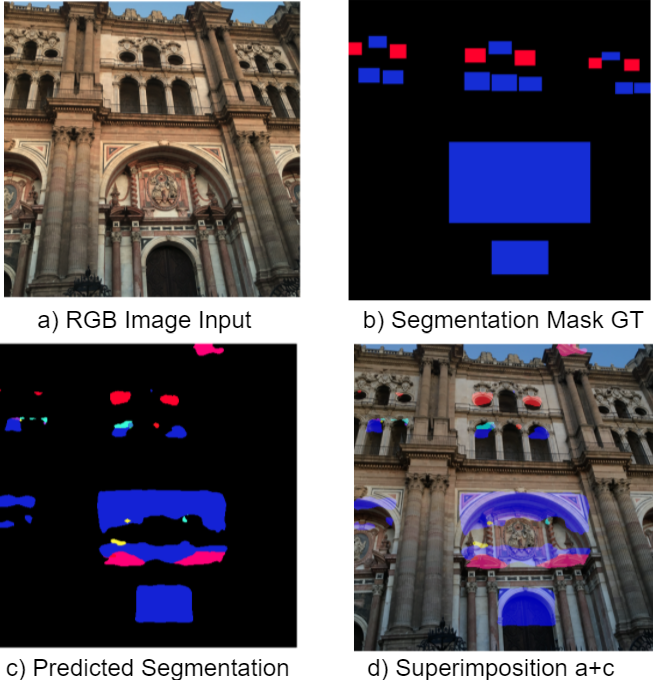} 
\caption{\adri{\textbf{Wrong segmentation, wrong prediction.}}. Example of results obtained using semantic segmentation. Image a) represents the RGB input data and Image b) represents the semantic segmentation masks of the Ground Truth, which the Latent Space Predictor must predict. Image c) represents the result of this segmentation and image d) is an overlay of the prediction on the input image, in order to see where the detected attributes are. The black pixels represent the background, the colored pixels represent architectural attributes. The red and blue attributes are missing several times and the yellow, pink and cyan where added while they do not exist in the GT. As the \textit{Latent Space Predictor} segments the image badly, the \textit{Transparent Classifier} is not able to correctly predict the class.}
\label{fig:example_3}
\end{figure*}
    \end{example}
    
\adri{\subsection{Counterfactual Explanations}
While an explanation refers to a description of the internal state or logic of an algorithm that leads to a decision, a counterfactual describes a dependency on external facts that led to that decision \cite{Wachter}. This allows to have an actionable element in the explanation, allowing to know what could have caused a change in the prediction \cite{mothilal2020explaining}. In short, a counterfactual is the answer to the question "If X had not occurred, Y would not have occurred.". It describes the smallest plausible change \cite{DelSer22} to the feature values that changes the prediction to a predefined output \cite{Verma}. Several methods exist to produce counterfactual explanations \cite{Cf1, Cf2, Cf3, Cf4, Cf5}. With \textit{Greybox XAI} framework, we propose to answer the question "Which object would have been predicted if a different object part had been detected?" to find the equivalent of a counterfactual example for a specific prediction. We do not propose to compute what would be the minimal change in the object prediction, or in the initial semantic segmentation, but rather to see what the final prediction would have been if the \textit{Explainable Latent Space} of detected object parts had been different. For a given attribute vector $z_{att,i}$ detected by the \textit{Latent Space Predictor}, the probability that the predicted class $y_i$ is the class $k$ is expressed as follows:
\begin{equation}
    p(y_i=k | z_{att,i} ; \theta_h) = \frac{\exp{(\theta_{h,j}^{\intercal} z_{att,i})}}{\sum_{j=1}^K(\theta_{h,j}^{\intercal} z_{att,i,j})}
\end{equation}
with $\theta_{h,j}^{\intercal}$ the weight matrix of the \textit{Transparent Classifier}. The attribute vector $z_{att,i}$ being composed of 0 and 1, each probability $y_i=k$ is written as an exponential fraction of a sum of weights. Thus, it is straightforward to calculate how much a change between a 1 and a 0 in the attribute vector $z_{att,i}$ would change in the prediction.
}

\section{Discussion}

Although we have shown that the \textit{Greybox XAI} framework, a compositional transparent model, is capable of achieving an image classification with a satisfying accuracy, this is largely caused by the quality of the attributes present in the dataset describing the set of classes. Without the presence of discriminative attributes in the dataset, it would not be possible to perform this kind of compositional classification. \adri{As an example, we can see on Figure \ref{fig:KG_Monumai} that the classes Baroque and Renaissance share 3 attributes: Rounded Arch, Lintelled Doorway and Porthole. Thus, if only these 3 attributes are detected, it is possible that the network is wrong. Figure \ref{fig:weight_linreg_monumai} gives the weights of these attributes for the Baroque and Renaissance classes: }
\begin{itemize}
    \item Rounded Arch (Baroque: 0.13, Renaissance: 0.01)
    \item Lintelled Doorway (Baroque: 0.07, Renaissance: 0.01)
    \item Porthole (Baroque: 0.1, Renaissance: 0.04)
\end{itemize}

\adri{Thus, the presence (i.e., as output by the (object-part) detector) of these 3 elements will cause the prediction "Baroque" when it could just as well be, according to the expert KG, a Renaissance monument. Further human evaluation may be required to inspect outputs of these types, e.g., verifying that indeed the highest probabilities detected object parts indeed correspond to the top-most probable detected components in that image, and that no other object element that entirely discriminates to predict a particular class exists.}

In addition, the \textit{Transparent Classifier} accuracy depends on the \textit{Latent Space Predictor} performances. Moreover, using a classification based on concrete features forsakes one of the advantages of neural networks, which is to use abstract features. Furthermore, this framework is only compatible with a classification task. Several usual Computer Vision tasks (segmentation, detection) cannot be explained by this framework. Therefore, there is no explainability progress for the \textit{Latent Space Predictor}. Finally, the properties of the attributes (position, dimension, numbers) are not used by the \textit{Transparent Classifier} resulting in a loss of information. 

However, the gain in explainability is important and the possibility to know exactly why a certain prediction happened is very useful. \adrien{Our framework allows inserting human knowledge, since the annotation of attributes in the image to be classified is indeed done by a human. Furthermore, we show that our framework can elicit explanations relating attributes to predicted labels, which can be offered in a human-readable form and serve for validating whether the explanations of the framework concur with his/her expert knowledge.} We hope this will motivate the research community to build datasets that inherently provide a greater granularity and hierarchy in the organisation of datasets, in order to move from the widely used (image, class) pair to triples (images, attributes, classes).

\adri{An important element to note about this model is its \textit{actionability}. As we have seen in the examples of the Figures \ref{fig:example_1}, \ref{fig:example_2} and \ref{fig:example_3}, having the possibility to visualize the explanations of the predictions made by the model allows to explore ways to improve its performances, by correcting its predictions, or the validity of the explanations, by correcting the semantic segmentation.}

\section{Conclusions and Future Work}

The contribution proposed in this article are:

\begin{itemize}
    \item A formalisation of what is a "good" explanation. We propose 3 criteria - objectivity, intrinsicality, validity - to assert that an explanation is "good" or not.
    \item The \textit{Greybox XAI}, a compositional framework for explainable classification. It is composed by a black box \textit{Latent Space Predictor} and a \textit{Transparent Classifier}.
\end{itemize}

We proved that this framework is transparent and produces "good" explanations, based on attributes segmented on the image. We tested it on 2 datasets and showed that it has SOTA results compared to compsotional models. Nevertheless, the accuracy is lower than for opaque end-to-end CNNs like ResNet101 or Transformers like DeiT-B. 

One of the reasons why accuracy is not at the level of models like DeiT-B is that the semantic segmentation performed by the Latent Space Predictor is not perfect. This means that the \textit{Transparent Classifier} takes as input a sometimes distorted or incomplete representation of the RGB image to classify. Since the vectorization of the segmentation map do not include a gradient, it is not possible to perform a direct backpropagation between the \textit{Transparent Classifier} and the \textit{Latent Space Predictor}. Future work could therefore involve finding a way to influence the semantic segmentation using the \textit{Transparent Classifier}. 
Also, the representation of the image in the \textit{Explainable Latent Space} causes a significant loss of information because the attributes are not enumerated and their relative dimensions and positions are not taken into account. Future work will involve the encoding of this information in the \textit{Explainable Latent Space}, in order to be able to explain a prediction based on the size and position of the objects.
Finally, the \textit{Transparent Classifier} is here a simple logistic regression. The use of Logic Tensor Networks \cite{donadello2018semantic} could allow to improve the model by using real logic beyond unique is-a relationships. \adri{Also, it would be interesting to test this framework on a new benchmark dataset specifically containing images that are labeled with position of objects, object groupings and relational concept such as those involved in solving the challenges of Kandinsky Patterns\cite{MULLER2021103546, HolzingerIQ}}

\adri{	
Finally, let's take look at the Renaissance line in the Figure \ref{fig:weight_linreg_monumai}. We see that the attributes are not very discriminative in favor of this class: the highest positive value is a weight of 0.14 for the attribute \textit{Serliana}. On the contrary, some attributes like \textit{Ogee Arch} or \textit{Column Salomnic} have a weight of -0.38 and -0.36, which implies that their presence leads to an absence of \textit{Renaissance} monument. Thus, it may be difficult to recognize a renaissance monument because if a Serliana is not detected the class will have difficulty in gaining the upper hand. It is therefore interesting to check whether this particular attribute is detected with good accuracy.
An attribute like \textit{Triangular Pediment} which has for biggest absolute weight a 0.03 is almost useless in the prediction. We can therefore give it less importance during the semantic segmentation.
Thus, the weight matrix could be used in the future to fine-tune the semantic segmentation so as to give importance to certain attributes in particular (the highest absolute value like \textit{Horseshoe Arch} or \textit{Ogee Arch}) and to neglect others.}

\section{Acknowledgement}

This research was funded by the French ANRT (Association Nationale Recherche Technologie — ANRT) industrial CIFRE PhD contract with SEGULA Technologies.
Díaz-Rodríguez is supported by Juan de la Cierva Incorporación grant IJC2019-039152-I funded by 
MCIN/AEI /10.13039/501100011033 by “ESF Investing in your future” and Google Research Scholar Program. 
J. Del Ser acknowledges support from the Department of Education of the Basque Government (Consolidated Research Group MATHMODE, IT1456-22).

\bibliographystyle{elsarticle-num}
\bibliography{refs}

\newpage

\end{document}